\documentclass[lettersize,journal]{IEEEtran}

\usepackage{acro}

\usepackage{algorithm} 
\usepackage{algorithmic} 

\usepackage{array} 
\usepackage{booktabs} 
\usepackage{blkarray}
\usepackage{multirow} 

\usepackage{subcaption} 

\usepackage{color} 
\usepackage[dvipsnames]{xcolor} 

\usepackage[utf8]{inputenc} 

\usepackage{enumitem} 

\usepackage{epsfig} 
\usepackage{float} 
\usepackage{graphicx} 

\usepackage{setspace} 
\usepackage{stfloats} 

\usepackage{amsmath} 
\usepackage{amssymb} 
\usepackage{amsfonts} 
\usepackage{amsthm} 

\usepackage{cite} 
\usepackage{hyperref} 

\usepackage{textcomp} 
\usepackage{lipsum} 
\usepackage{soul} 

\usepackage{ifpdf} 
\usepackage{url} 
\usepackage{verbatim} 
\usepackage{optidef}
\usepackage{bm} 

\usepackage[table]{xcolor}
\definecolor{mygray}{rgb}{0.9,0.9,0.9} 
\usepackage{threeparttable}
\usepackage{nicematrix}

\usepackage{balance}
\usepackage{graphicx}  
\usepackage{subcaption} 
\usepackage{caption} 

\usepackage{xcolor} 

\DeclareAcronym{AFAS}{short = AFAS,
	long = Automated Fabric Alignement System
}

\DeclareAcronym{FAS}{short = FAS,
	long =  Fabric Alignement System
}

\DeclareAcronym{CAD}{short = CAD,
	long = Computer Aided Design
	}

\DeclareAcronym{DoF}{short = DoF,
	long = Degrees of Freedom
	}
	
\DeclareAcronym{PCR}{short = PCR,
	long = Point Cloud Registration
}

\DeclareAcronym{ICP}{short = ICP,
	long = Iterative Closest Point
}

\DeclareAcronym{DFAS}{
	short = DFAS ,
	long = Differential Fabric Alignment Sewing
}

\DeclareAcronym{DFFSC}{
	short = DFFSC ,
	long = Differential Fabric Feeding Speed Control
}

\DeclareAcronym{DFRoSS}{
	short = DFRoSS ,
	long = Differential Fabric RoSS
}

\DeclareAcronym{GLW}{short = GLW,
	long = Global Local Weighted
	}

\DeclareAcronym{IK}{short = IK,
	long = Inverse Kinematics
	}

\DeclareAcronym{RoSS}{
	short = RoSS ,
	long = Robotic Sewing System
}

\DeclareAcronym{FFRoSS}{
	short = FFRoSS ,
	long = Fixed Fabric Robotic Sewing System
}

\DeclareAcronym{NFRoSS}{
	short = NFRoSS ,
	long = Non-fixed Fabric Robotic Sewing System
}

\DeclareAcronym{FST}{
	short = FST ,
	long = Fabric Stabilizing Tool
}

\newcommand{\ieeecopyrighttext}{%
  \footnotesize
  \textcopyright{}~2026 IEEE. Personal use of this material is permitted.
  Permission from IEEE must be obtained for all other uses, in any current or future
  media, including reprinting/republishing this material for advertising or promotional
  purposes, creating new collective works, for resale or redistribution to servers or
  lists, or reuse of any copyrighted component of this work in other works.}

\makeatletter
\def\@IEEEpubidpullup{5.5\baselineskip}
\makeatother

\newcommand{\setupieeecopyright}{%
  \IEEEpubid{%
    \makebox[\textwidth][l]{%
      \parbox[t]{0.48\textwidth}{\ieeecopyrighttext}}}}

\newcommand{\publicationnotice}{%
  \vspace{-0.75em}
  \begin{center}
  \small
  This paper has been published in \emph{IEEE Transactions on Automation Science and Engineering},
  vol.~23, pp.~13391--13406, July 2026.
  The definitive version is available at
  \url{https://doi.org/10.1109/TASE.2026.3713128}.
  An interactive project page with videos and results is available at
  \url{https://bhattner143.github.io/rfas-glwicp.github.io/}.
  \end{center}
  \vspace{0.25em}}

\def\BibTeX{{\rm B\kern-.05em{\sc i\kern-.025em b}\kern-.08em
    T\kern-.1667em\lower.7ex\hbox{E}\kern-.125emX}}

\begin{document}

\title{Robotic Fabric Alignment System for Sewing Using\\ Global Local Weighted ICP}
\author{Wenbo Dong~\IEEEmembership{Graduate Student Member,~IEEE}, Dipankar Bhattacharya~\IEEEmembership{Member,~IEEE}, Kai Tang~\IEEEmembership{Graduate Student Member,~IEEE}, Akinari Kobayashi,~\IEEEmembership{Member,~IEEE}, Fuyuki Tokuda,~\IEEEmembership{Member,~IEEE}, Akira Seino,~\IEEEmembership{Member,~IEEE}, \\Norman C. Tien,~\IEEEmembership{Senior Member,~IEEE}, and Kazuhiro Kosuge,~\IEEEmembership{Life Fellow,~IEEE}

\thanks{W. Dong and K. Tang are with the JC STEM Lab of Robotics for Soft Materials, Department of Electrical and Computer Engineering, Faculty of Engineering, The University of Hong Kong, Hong Kong SAR, China. (e-mail: dongwbo@connect.hku.hk; tangkai@eee.hku.hk).}
\thanks{Dipankar Bhattacharya is with the Dyson School of Design Engineering, Imperial College London, London, United Kingdom. (e-mail:d.bhattacharya1@imperial.ac.uk)}
\thanks{A. Kobayashi was with the JC STEM Lab of Robotics for Soft Materials, Department of Electrical and Computer Engineering, Faculty of Engineering, The University of Hong Kong, Hong Kong SAR, China.(e-mail: akinari.kobayashi.hk@gmail.com).}
\thanks{F. Tokuda is with the Unprecedented-scale Data Analytics Center, Tohoku University, Sendai 980-0845, Japan, and also with the Graduate School of Information Sciences, Tohoku University, Sendai 980-0845, Japan. (e-mail: fuyuki.tokuda.b3@tohoku.ac.jp).}
\thanks{A. Seino is with Faculty of Symbiotic Systems Science, Fukushima University, Fukushima 960-1296, Japan, (e-mail: akira\_seino@sss.fukushima-u.ac.jp).}
\thanks{N. C. Tien is with the Department of Electrical and Computer Engineering, Faculty of Engineering, The University of Hong Kong, Hong Kong SAR, China. (e-mail: nctien@hku.hk).}
\thanks{K. Kosuge is with the Department of Mechanical Engineering, City University of Hong Kong, Hong Kong SAR, China. (e-mail: kkosuge@cityu.edu.hk).}
}

\setupieeecopyright
\maketitle
\IEEEpubidadjcol
\publicationnotice
\begin{abstract}
Accurate fabric alignment is a critical step that must be performed before sewing. 
This paper presents a novel automated fabric alignment system. The system estimates the poses of top and bottom fabric panels—lying flat and wrinkle-free in arbitrary positions—using a new Global Local Weighted Iterative Closest Point (GLW-ICP) method. The system then manipulates the top panel to achieve precise alignment at both edges and sewing lines.
Unlike conventional approaches, GLW-ICP robustly aligns both global edges and local sewing lines by globally aligning fabric edge points and locally aligning sewing line points to their corresponding CAD model points, while removing unmatched points in occluded regions. 
Real-world experiments with various fabric shapes show that the system consistently achieves millimeter-level alignment accuracy under both occlusion and non-occlusion conditions, demonstrating its effectiveness and suitability for automated fabric alignment in practical scenarios.

\end{abstract}

\def\abstractname{Note to Practitioners}
\begin{abstract}
Fabric panel alignment before sewing is a time-consuming and skill-dependent task in garment production. Misaligned edges or sewing lines can lead to defects, rework, and production delays. This work presents a novel robotic system that automates the alignment of wrinkle-free fabric panels, even when portions of the panel are occluded by the manipulator. The system uses a novel Global Local Weighted Iterative Closest Point (GLW-ICP) method, which separately aligns overall panel edges and local sewing lines to a digital CAD model while ignoring unreliable points from occluded regions. A roller-based end-effector then picks up, re-positions, and releases the top panel to achieve precise alignment with the bottom panel. This method achieves millimeter-level accuracy across various garment components under both unoccluded and partially occluded views. This reduces operator dependency, improves consistency, and shortens preparation time. The approach is readily applicable to a wide range of garment components and can be adapted to various production settings. These capabilities open opportunities for end-to-end automation in apparel manufacturing, from panel preparation to stitching, further enhancing productivity and quality control.    
\end{abstract}


\section{Introduction}

{\color{black}
\IEEEPARstart{F}{abric} alignment for sewing is a critical step in which two stacked fabric pieces—the top and bottom panels—must be aligned before stitching. However, fabric alignment is labor-intensive, accounting for approximately 75\% of the total sewing operation time~\cite{ngan2011automated, jana2015sewing}. To facilitate accurate fabric alignment, panels are often held flat using suction tables to minimize the wrinkles and enable repeatable positioning~\cite{kuvzel2022vacuum}, while computer-aided design (CAD)-based pattern data are used to ensure geometric consistency across layers~\cite{atalie2020application}. Even with these standard aids, slight misalignment inevitably occurs due to local fabric deformation during handling. This misalignment directly impacts the final product’s shape and quality. 

\begin{figure}[t!]
	\centering
	\includegraphics[width=1\columnwidth]{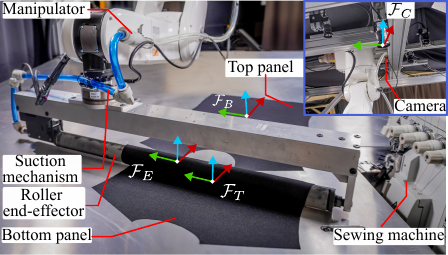}
	\caption{Proposed automatic fabric alignment system.}
	\label{fig:alignment_system} 
    \vspace{-5mm}
\end{figure}

The fabric alignment is difficult to automate and is typically performed manually in traditional or semi-automatic systems. This makes the process time-consuming and highly dependent on the operator’s skill~\cite{meng2022automatic}. Although fabric alignment systems for sewing were proposed in~\cite{ku2023automated, kim2023robotic}, both papers assume that fabrics are pre-loaded onto the template manually. 
To the best of the authors' knowledge, no system exists that can automatically align and stack the two fabric pieces.

In this paper, we propose a robotic fabric alignment system that blackuces manual intervention before sewing. The proposed fabric alignment system has two main functions: fabric pose estimation and fabric manipulation. Fabric pose estimation determines the position and orientation of the top and bottom panels based on the CAD models. The CAD model of each panel includes the geometry of the cut edges as global information and the desiblack seams along designated edges for seaming as local information. 

While the cut edge geometry is a visible global feature commonly used for alignment, relying on it alone often leads to seam displacement. Fabric cutting errors, caused by misaligned layers during multi-layer cutting, blunt tools~\cite{vilumsone2018industrial}, or effects such as shrinkage, fraying, and tolerance-induced deviations, cause the cut edge of the real fabric to deviate from the intended CAD geometry. Therefore, the cut edge is often an insufficient reference for precise alignment and sewing.

In standard manual garment construction, sewing lines are introduced for precise fabric seaming. Unlike cut edges, sewing lines represent the true intended seam path. Therefore, aligning the sewing lines of the top and bottom panels is essential for garment quality. Human operators can align the fabrics using local seam information since they know the geometric relationship between each seam and the global cut edges. Thus, to automate the fabric alignment, pose estimation using both global and local information is necessary.

Pose estimation is typically formulated as a registration problem between the observed fabric geometry and a known CAD model~\cite{besl1992method}. While \ac{ICP} method is commonly used to minimize spatial discrepancies in this context, directly applying them presents three major challenges. First, standard ICP treats all correspondences uniformly and cannot distinguish between geometrically distinct features. This causes the alignment to be affected by erroneous edge points resulting from cutting errors, which can misplace the sewing line~\cite{4982554,1430845,bearee2011innovative}. Second, ICP performs poorly with partially occluded fabrics. While variants such as Trimmed-ICP~\cite{chetverikov2002trimmed} and Sparse-ICP~\cite{bouaziz2013sparse} improve robustness by discarding high-error correspondences or using sparsity constraints, they depend heavily on parameter tuning or increase computational complexity. Third, existing ICP methods are sensitive to noise, poor initial alignment, and outliers~\cite{rusinkiewicz2001efficient}, including erroneous points arising from fabric cutting errors.

This paper presents an automated fabric alignment system (Fig.\ref{fig:alignment_system}) that estimates the poses of top and bottom fabric panels—lying flat and wrinkle-free—using a novel ICP method that incorporates both global and local information. The system then picks up the top panel with a roller end-effector, re-estimates its pose under partial occlusion, and precisely aligns it with the bottom panel.

The main contributions of this paper are as follows:}
\begin{itemize}
    {\color{black} 
    \item A novel robotic fabric alignment system is developed, that integrates fabric pose estimation under occlusion and robotic manipulation. To the best of the authors' knowledge, this is the first system capable of aligning the edges and sewing lines of partially visible fabrics.

    \item \ac{GLW}-\ac{ICP} is proposed for pose estimation of partially visible fabrics, which introduces a dual, feature-aware weighting mechanism that separately aligns global edge points and local sewing line points to their corresponding CAD model points.
  }

  \item Experiments using different shapes of fabric demonstrate that the system achieves high alignment accuracy under partial occlusion and non-occlusion conditions.
    
\end{itemize}

The rest of this paper is organized as follows.
Sec.~\ref{sec:related_work} reviews related work on fabric alignment and pose estimation.
Sec.~\ref{sc:problem_formulation_and_outline}, and~\ref{sec:glwicp_pose_est} introduce the automated fabric alignment system, and the GLW-ICP, respectively. 
Sec.~\ref{sec:system_operation} presents the system workflow and implementation.
Sec.~\ref{sec:pose_est_and_fabric} shows experimental results, and Sec.~\ref{sec:conclusion} concludes with future directions.

\section{Related works} \label{sec:related_work}

\subsection{Existing fabric alignment systems}
Most existing fabric alignment systems are semi-automatic, with tasks like fabric manipulation automated, while fabric pose estimation still requires manual intervention. 
Existing commercial systems, such as pattern sewers and pocket setters, often use rigid fixtures or templates to constrain and align fabrics during sewing~\cite{ku2023automated}. 
Recent studies have explored various robotic systems to facilitate fabric handling. 
Parker et al.~\cite{parker1983robotic} developed a pin-based end-effector system for isolating single layers, but it is only effective for flat, simple shapes.
Manabe et al.~\cite{9419523} introduced a roller-hand mechanism for single-sheet separation using frictional interaction. 
Tajima et al.~\cite{9551585} developed a dual-arm robotic system with steerable rollers and visual feedback for fabric feeding and straight-line sewing. 
Yamazaki et al.~\cite{9345958} proposed a compact brush-based end-effector for pick-and-release of stacked cotton and woven fabrics. 
Ogura et al.~\cite{ogura2022automation} designed a robotic system combining a guide-set, a shape-conforming gripper, and a pre-shaped mold to align and fix irregularly shaped fabric parts.
Such systems, while effective in structured settings, are not fully automated and lack adaptability to shapes like collar and shirts.

\subsection{Existing pose estimation methods}\label{sec:existing_pose_estimation}

\subsubsection*{Vision-based methods}
Recent studies have investigated vision-based approaches for fabric pose estimation, focusing on extracting alignment features such as fabric edges, contours, and shapes. 
Schrimpf et al.~\cite{schrimpf2012experiments, schrimpf2014velocity} proposed a robotic system that utilizes optical edge sensors for fabric position estimation and seam tracking, as well as force feedback for fabric motion synchronization.
Tokuda et al.~\cite{10715572} developed a fixture-free 2D sewing system using a dual-arm manipulator and visual feedback control to track fabric contours and guide stitching paths. 
Torgerson et al.~\cite{torgerson1988vision} presented a vision-guided robotic system that analyzed fabric boundary shapes and computed seam paths for fabric manipulation. 
Ku et al.~\cite{10197511} proposed a machine vision-based sewing system that segments sewing lines under varying lighting conditions using deep learning. 
While these methods can handle fabrics, they often rely on clear visibility, so occlusion or distortion from folds, overlaps, or complex textures can reduce alignment accuracy.

\subsubsection*{\ac{PCR} methods}
Unlike vision-based approaches, PCR methods align 3D point sets by estimating spatial transformations. 
Under occlusion, Dang et al.~\cite{dang20203d} proposed a deep learning-based PCR framework that estimates the 6D pose of self-occluded objects in cluttered 3D scenes. 
{
\color{black}
However, its reliance on learned features limits generalization to unseen fabrics, and the rigidity assumption fails for deformable fabrics.
}
Ma et al.~\cite{ma2013robust} proposed a nonrigid point set registration method that robustly estimates transformations between point sets containing noise and outliers.
{
\color{black}
However, it assumes complete visibility and dense sampling, which do not hold in fabric alignment with partial occlusion and sparse geometric features.
}
Campbell et al.~\cite{campbell2015adaptive} proposed a sparse Gaussian mixture model-based representation that improves robustness to occlusion and missing data by selecting a subset of representative points that best describe the point-set shape, with adaptive weights that vary according to each point's contribution to the alignment.
{
\color{black}
However, this method relies on dense, uniformly distributed points to build a reliable surface model. With only sparse edge or sewing-line points, the Gaussian mixtures become ill-conditioned, and leading to inaccurate registration. Its iterative probabilistic optimization is also computationally expensive, limiting real-time fabric pose estimation.
}

\subsubsection*{ICP methods}
Various ICP variants of PCR have been developed for different application scenarios. 
{\color{black}
Standard ICP iteratively minimizes point‑to‑point distances but is sensitive to noise, partial occlusion, and poor initial alignment.
Trimmed-ICP~\cite{chetverikov2002trimmed} improves robustness by discarding high‑error correspondences, making it effective under noisy or partially occluded conditions, but its performance depends heavily on selecting an appropriate trimming ratio. 
Sparse-ICP~\cite{bouaziz2013sparse}  uses sparsity constraints to suppress outliers and enhance stability in cluttered or incomplete data, but its iterative optimization increases computational cost and complexity.
Probabilistic methods such as Coherent Point Drift (CPD)~\cite{myronenko2010point} enhance robustness to noise and partial data by modeling correspondences as distributions, though they are computationally expensive and rely on good initialization.
Fast and Robust ICP (FRICP)~\cite{9336308} accelerates convergence and improves resistance to outliers using advanced optimization and robust error metrics, but its implementation is more complex and still sensitive in misalignment cases.
Point Cloud Registration Based on Kendall Shape Space (KSS‑ICP)~\cite{10061449} aligns shapes in a normalized Kendall shape space, achieving invariance to scale, rotation, and translation, but it requires extensive preprocessing and performs poorly when point clouds are occluded or structurally inconsistent.

To handle noise and outliers, AW-RICP~\cite{guo2022adaptive} learns a single weight vector that adaptively selects sparse neighbors and assigns smaller weights to point pairs with larger registration errors.
While this improves robustness, the unified scheme treats all correspondences uniformly and cannot distinguish between geometrically distinct features.
Overall, compared to existing ICP variants, GLW-ICP is specifically designed for fabric alignment: it uses dual, feature-aware weights to prioritize reliable sewing-line geometry over error-prone cut edges, while robustly handling partial occlusion and measurement noise common in robotic sewing systems.
}
\section{Proposed system outline}\label{sc:problem_formulation_and_outline}

\begin{figure*}[t!]
    \centering
    \includegraphics[width=\textwidth]{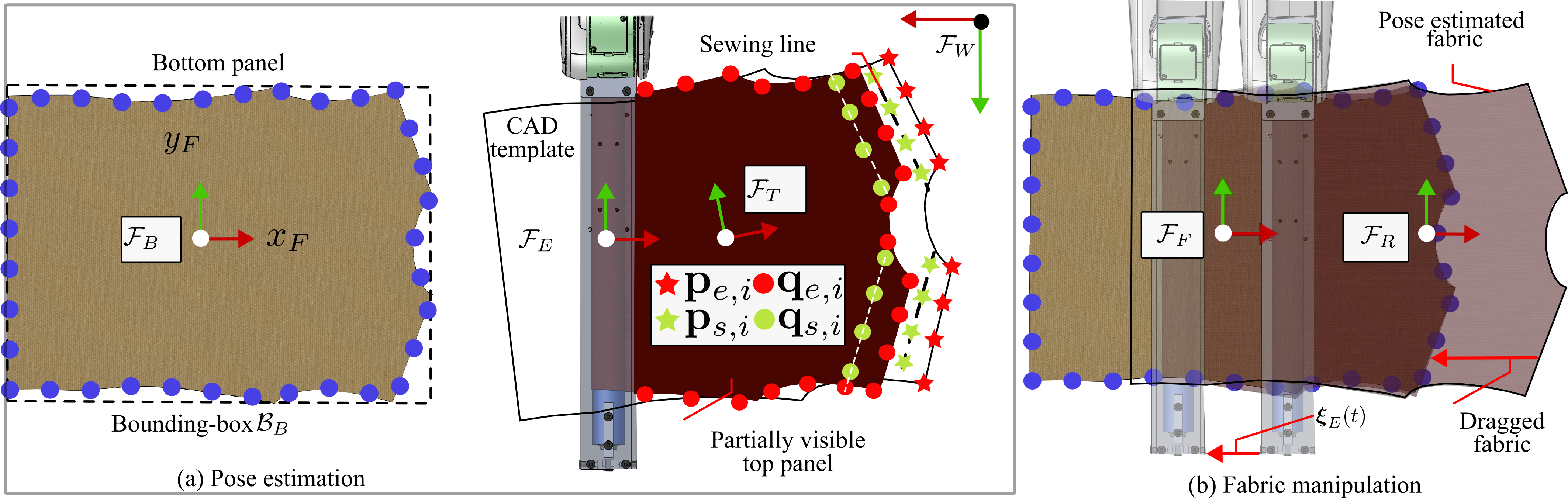} 
    \caption{Proposed fabric alignment process.}
    \label{fig:icp_formulation} 
    \vspace{-3mm}
\end{figure*}

This section presents the hardware system and outlines the overall proposed automated fabric alignment process.

\subsection{Proposed automated fabric alignment process}
\label{sec:automated_alignment_process}
The proposed fabric alignment process, as shown in Fig. \ref{fig:icp_formulation}, is described as follows.
1) Given two fabric panels—the \textit{top panel} and the \textit{bottom panel}—initially laid flat and wrinkle-free (Fig.~\ref{fig:icp_formulation}(a)) on a workbench and arbitrarily positioned in the camera view, the automated alignment process begins by capturing RGB images of both panels and extracting edge and sewing line points using camera intrinsic and extrinsic parameters.
{\color{black}
    Note that the top and bottom panels do not need to share the same CAD geometry. Each panel is registered independently to its own CAD model. This allows the system to handle panels with different shapes, as demonstrated in the shirt experiment (Sec.~\ref{sec:dataset}).}
2) The initial positions and orientations of the panels are determined by applying the proposed \ac{GLW}-\ac{ICP} algorithm to the observed and CAD model points. Once both panels are accurately located, the \textit{grasping points} on the top panel and \textit{target points} on the bottom panel are identified.
3) Using the known grasping point, the top panel is fully wrappped onto a roller-based end effector and moved to an \textit{offset target location} above the bottom panel's sewing line.
4) The top panel is partially released onto the plane, after which the GLW-ICP algorithm is reapplied to re-estimate the fabric pose from the now partial view (Fig.~\ref{fig:icp_formulation}(a)).
5) Using the known target points on the bottom panel, the top panel is gradually dragged to align its sewing line with that of the bottom panel (Fig.~\ref{fig:icp_formulation}(b)).
6) Once alignment is complete, the final release is performed to finalise the fabric alignment.

\subsection{System description}
\label{sec:sys_description}
To accomplish automated alignment (Sec.~\ref{sec:automated_alignment_process}), the system integrates the following primary hardware components:
\subsubsection{Manipulator and Roller End-Effector}
A 6-\ac{DoF} robotic arm (Denso VS-087) is mounted on an aluminum frame and controlled by servo controller (Denso RC8A) via EtherCAT. At the arm’s end, a force/torque sensor (ATI Axia80-M8) and a roller end-effector are installed, which features a motor-driven roller with a suction port for fabric pick-up, and a servo motor (Yaskawa Sigma-7 SGM7M-A3A3AA1) with rotary encoder for precise roller rotation.
\subsubsection{3D Vision system}
The vision system uses a Basler acA1920-155uc camera (1920×1280 @ 164 fps) with a 12mm lens (C11-1220-12M) for accurate fabric feature detection.
\subsubsection{Control workstation}
The control workstation includes 64 GB RAM, an Intel Xeon W-2295 CPU, and an NVIDIA GeForce RTX 4090 GPU. It operates on Windows OS and TenAsys INtime 6 OS for robotic control. 

\subsection{Coordinate system and pose representation}
\label{sec:coordinate_system}
To coordinate the alignment system hardware and fabrics, the following coordinate system is established (Fig.~\ref{fig:icp_formulation}). 
The world frame [\(\mathcal{F}_W\), origin \(\mathcal{O}_W\{x_W, y_W, z_W\}\)] is fixed on the platform desktop. 
The end-effector frame [\(\mathcal{F}_E\), origin \(\mathcal{O}_E\{x_E, y_E, z_E\}\)] is located at the center of the roller of the end-effector.
The camera frame [\(\mathcal{F}_C\), origin \(\mathcal{O}_C\{x_C, y_C, z_C\}\)] is attached to the camera, with \(y_C\) parallel to \(y_W\) and \(z_C\) aligned with \(z_W\). 
Throughout, a left-side superscript indicates the reference frame of each quantity.

For notational convenience, let \(i \in \{T, B\}\) denote the panel index, with \(T\) for the top panel and \(B\) for the bottom panel. Accordingly, \(\mathcal{F}_i\) and \(\mathcal{B}_i\) denote coordinate frame and bounding box for panel \(i\), respectively. 
The bounding box is constructed as the minimal axis-aligned 3D box that encloses the point cloud of the panel \(i\).
Each panel coordinate system \(\mathcal{F}_i\), origin \(\mathcal{O}_i\{x_i, y_i, z_i\}\) is centered and aligned with its bounding box \(\mathcal{B}_i\).
The axes of each frame \(\mathcal{F}_i\) are set by the geometry of its bounding box \(\mathcal{B}_i\), the \(x_i\)-axis aligns with the longer side, and the \(y_i\)-axis with the shorter side.
The pose of panel \(i\) in the world frame \(\mathcal{F}_W\) is specified by its position \([x_i, y_i, z_i]^\top\) and orientation \([\theta_{i,x}, \theta_{i,y}, \theta_{i,z}]^\top\), where the orientation components are Euler angles about the \(x\)-, \(y\)-, and \(z\)-axes of \(\mathcal{F}_W\). Thus, the pose can be represented as a generalized coordinate vector,
\(
    \boldsymbol{\xi}_i = [x_i, y_i, z_i, \theta_{i,x}, \theta_{i,y}, \theta_{i,z}]^\top,
\) where \(i \in \{T, B\}.\)
Given the pose parameters, the transformation from the local frame \(\mathcal{F}_i\) to the world frame \(\mathcal{F}_W\) is represented by homogeneous transformation matrix
\begin{equation}
\label{eq:transformation_W_i}
 \textstyle
    {}^W\mathbf{T}_i = \begin{bmatrix} {}^W\mathbf{R}_i & {}^W\mathbf{t}_i \\ \mathbf{0}^\top & 1 \end{bmatrix},
\end{equation}
where \({}^W\mathbf{R}_i \in \mathbb{R}^{3 \times 3}\) is a rotation matrix (\({}^W\mathbf{R}_i^\top {}^W\mathbf{R}_i = \mathbf{I}\), \(\det({}^W\mathbf{R}_i) = 1\)), and \({}^W\mathbf{t}_i \in \mathbb{R}^3\) is the translation vector, both expressing the orientation and position of panel \(i\) in \(\mathcal{F}_W\).
The generalized pose vector can be recovered from the transformation matrix as
\begin{equation}
    \label{eq:generalized_pose}
    \boldsymbol{\xi}_i = \text{ToGeneralized}\left({}^W\mathbf{T}_i\right),
    \quad i \in \{T, B\},
\end{equation}
where \(\text{ToGeneralized}(\cdot)\) extracts the position and orientation from the transformation matrix.

\subsection{Fabric pose estimation overview}
\label{sec:fabric_pose_est}
At time \( t = 0 \), the manipulator has partially released the top panel fabric, making it only partially visible. The initial pose \( \boldsymbol{\xi}_T(0) \), as defined in~\eqref{eq:generalized_pose} with \( i = T \), is unknown and must be estimated.
To facilitate this estimation, two sets of points are generated from the RGB image of the top panel fabric, as shown in Fig.~\ref{fig:icp_formulation}, and represented in the world frame \( \mathcal{F}_W \). 
The first set, \textit{global edge target points}, consists of \( n_e \) edge points (red circles) and is represented by the matrix \( \mathbf{Q}_e = [\mathbf{q}_{e,1}, \dots, \mathbf{q}_{e,n_e}] \in \mathbb{R}^{3 \times n_e} \), where each \( \mathbf{q}_{e,i} \in \mathbb{R}^3 \).
The second set, \textit{local sewing target points}, contains \( n_s \) points along the sewing line (green circles), represented by \( \mathbf{Q}_s = [\mathbf{q}_{s,1}, \dots, \mathbf{q}_{s,n_s}] \in \mathbb{R}^{3 \times n_s} \), with \( \mathbf{q}_{s,i} \in \mathbb{R}^3 \).

Given that the pose of CAD model in \(\mathcal{F}_W\) is specified by known homogeneous transformation matrix \( {}^W\mathbf{T}_{CAD} \), two sets of key points are defined on CAD model. 
The first set, \textit{global edge source points}, consists of \( m_e \) edge points (red stars) represented by the matrix \( \mathbf{P}_e = [\mathbf{p}_{e,1}, \dots, \mathbf{p}_{e,m_e}] \in \mathbb{R}^{3 \times m_e} \), where each \( \mathbf{p}_{e,i} \in \mathbb{R}^3 \).
The second set, \textit{local sewing source points}, contains \( m_s \) the sewing line points (green stars), represented by \( \mathbf{P}_s = [\mathbf{p}_{s,1}, \dots, \mathbf{p}_{s,m_s}] \in \mathbb{R}^{3 \times m_s} \), with \( \mathbf{p}_{s,i} \in \mathbb{R}^3 \).

The objective of pose estimation is to determine the \textit{optimal rigid transformation} as in~\eqref{eq:transformation_W_i}, where $\mathbf{R}^\star \in \mathbb{R}^{3 \times 3}$ and $\mathbf{t}^\star \in \mathbb{R}^3$ denote the optimal rotation matrix and translation vector, respectively. This transformation, computed using GLW-ICP (Sec.~\ref{sec:glwicp_pose_est}), aligns the global edge and local sewing source points of the CAD model with their corresponding target points from the top panel fabric. Alignment is achieved when the following errors, evaluated at $\mathbf{R}^\star$ and $\mathbf{t}^\star$, are minimized
\begin{align}
    \label{eq:E_e}
    \textstyle
    g_e(c, \mathbf{W}_e, \mathbf{R}^\star, \mathbf{t}^\star) &= \left\| \mathbf{E}_e(c, \mathbf{R}^\star, \mathbf{t}^\star) \right\|_{F, \mathbf{W}_e}^2, \\
    \label{eq:E_s}
    \textstyle
    g_s(c, \mathbf{W}_s, \mathbf{R}^\star, \mathbf{t}^\star) &= \left\| \mathbf{E}_s(c, \mathbf{R}^\star, \mathbf{t}^\star) \right\|_{F, \mathbf{W}_s}^2,
\end{align}
where the \textit{global alignment term} \(g_e(\cdot)\) aligns the CAD model edge points \(\mathbf{P}_e\) to the corresponding fabric edge points \(\mathbf{Q}_{e,c}\in\mathbb{R}^{3\times n_e}\), while the \textit{local alignment term} \(g_s(\cdot)\) aligns the CAD sewing line points \(\mathbf{P}_s\) to the fabric sewing line points \(\mathbf{Q}_{s,c}\in\mathbb{R}^{3\times n_s}\).
In these expressions, \( \mathbf{E}_e(\cdot) \) is the \textit{global alignment error} for the edge points, measuring the pose estimation accuracy between the edge in CAD model and the fabric edge; \( \mathbf{E}_s(\cdot) \) is the \textit{local alignment error} for the sewing line points, reflecting the alignment accuracy of the sewing line; and \( c \) denotes the correspondence mapping between the CAD model points and the fabric points. 
The weighted Frobenius norms and error matrices are given by
\begin{align}
    \label{eq:E_i_frob_norm}
    &   \left\| \mathbf{E}_i (\cdot)\right\|_{F, \mathbf{W}_i}^2
    = \operatorname{tr}\left( \mathbf{E}_i(\cdot)\, \mathbf{W}_i\, \mathbf{E}_i(\cdot)^\top \right),\quad i \in \{e, s\}, \\
    &\text{where}\quad\mathbf{E}_i(\cdot) = \mathbf{R}^\star \mathbf{P}_i + \mathbf{t}^\star \mathbf{1}_{n_i}^\top - \mathbf{Q}_{i,c},
\end{align}
where \(\operatorname{tr}(\cdot)\) denotes the trace operator, and the diagonal weight matrices \(\mathbf{W}_i = \operatorname{diag}(\mathbf{w}_i) \in \mathbb{R}^{n_i \times n_i}\) assign importance to individual points in each set, where \(\mathbf{w}_i = [w_{i,1}, \dots, w_{i,n_i}]^\top \in \mathbb{R}^{n_i}\) for \(i \in \{e, s\}\). Given the optimal \(\mathbf{R}^\star\) and \(\mathbf{t}^\star\), the pose of the top panel fabric in \(\mathcal{F}_W\) can be determined using~\eqref{eq:generalized_pose} as
\begin{align}
    \label{eq:generalized_pose_rear}
    \textstyle
    \boldsymbol{\xi}_T(0) = \text{ToGeneralized}\left({}^W\mathbf{T}_{T}(0)\right),
\end{align}
where the initial top panel fabric transformation with respect to \(\mathcal{F}_W\), \({}^W\mathbf{T}_{T}\), is given by
\begin{align}
\label{eq:T_T}
    {}^W\mathbf{T}_{T}={}^W\mathbf{T}_{CAD}(\mathbf{T}^\star)^{-1}.
\end{align}

\subsection{Fabric manipulation for alignment overview}

With the top panel fabric pose \( \boldsymbol{\xi}_T(0) \) from~\eqref{eq:generalized_pose} computed by the GLW-ICP algorithm, the goal is to move the top panel so its pose and sewing line match those of the wrinkle-free bottom panel \( \boldsymbol{\xi}_B \). This is achieved using a roller end-effector, which aligns the partially released top panel fabric with the bottom panel before fully releasing it (Fig. \ref{fig:icp_formulation}). 
To initiate alignment at \( t = 0 \), the initial end-effector pose—corresponding to partial fabric release—must be determined. This pose can be directly computed in the world frame \( \mathcal{F}_W \), since the end-effector's pose relative to the top panel (\({}^T\mathbf{T}_E\)) is known and assumed constant during motion. Specifically,
\begin{equation}
\label{eq:T_W_E}
{}^W\mathbf{T}_E(0) = {}^W\mathbf{T}_T(0) \cdot {}^T\mathbf{T}_E,
\end{equation}
where \( {}^W\mathbf{T}_T(0) \) is given by~\eqref{eq:T_T}. Using~\eqref{eq:T_W_E}, the corresponding generalized pose of the end-effector in \(\mathcal{F}_W\) is
\begin{equation}
    \label{eq:ee_initial_pose}
    \boldsymbol{\xi}_E(0) = \text{ToGeneralized}\left({}^W\mathbf{T}_T(0) \cdot {}^T\mathbf{T}_E\right).
\end{equation}

At the final time \( t_f \), successful fabric alignment yields 
\begin{equation}
\label{eq:alignment_matrix}
{}^W\mathbf{T}_T(t_f) = {}^W\mathbf{T}_B 
\quad \text{or equivalently,} \quad 
\boldsymbol{\xi}_T(t_f) = \boldsymbol{\xi}_B
\end{equation}
ensuring the top panel matches the bottom panel in both position and orientation.
Hence, from \eqref{eq:alignment_matrix} the end-effector's final pose can be written as
\begin{equation}
    \label{eq:ee_final_pose}
    \boldsymbol{\xi}_E(t_f) = \text{ToGeneralized}\left({}^W\mathbf{T}_B \cdot {}^T\mathbf{T}_E\right),
\end{equation}
where it has been assumed that the top panel fabric remains wrinkle-free during the manipulation, ensuring that \({}^T\mathbf{T}_E\) remains constant.

Finally, to manipulate the fabric from its initial state \(\boldsymbol{\xi}_E(0)\) to the target state \({}^W\mathbf{T}_B\), a fifth-order polynomial trajectory is used. This ensures smooth, continuous motion and maintains the wrinkle-free condition of the fabric.

\newtheorem{remark}{Remark}

\section{Pose estimation by GLW-ICP} 
\label{sec:glwicp_pose_est}

This section discusses the formulation and solution of the GLW-ICP optimization problem. For consistent spatial referencing, all edge and sewing line points are defined with respect to the world frame (\(\mathcal{F}_W\)) introduced in Sec.~\ref{sec:coordinate_system}.

\subsection{GLW-ICP optimization problem formulation}
\label{sec:glwicp_opt_problem_form}

To determine the optimal rigid transformation $\mathbf{T} = [\mathbf{R}|\mathbf{t}]$, 
an optimization problem is formulated that minimizes the weighted alignment error 
between source and target points. 
The objective function combines four components: 
\textit{alignment terms} for global edges and local sewing lines, derived from~\eqref{eq:E_e} and \eqref{eq:E_s}, and \textit{regularization terms} that stabilize 
the optimization by penalizing large weight values for each. 
This yields
\begin{align}
    \label{eq:min_g}
    \textstyle
    &\mathbf{T}^\star =[\mathbf{R}^{\star}|\mathbf{t}^{\star}]=\arg\min_{
        \substack{
            c,\, \mathbf{W}_e,\, \mathbf{W}_s, \\
            \mathbf{R},\, \mathbf{t}
        }
    }\, g(c, \mathbf{W}_e, \mathbf{W}_s, \mathbf{R}, \mathbf{t}), \\
    \textstyle
    &g = \underbrace{\textstyle\sum_{i \in \{e, s\}} g_i(\cdot)}_{\text{alignment}} 
    + \alpha\, \underbrace{g_r(\mathbf{W}_e, \mathbf{W}_s)}_{\text{regularization}}, \label{eq:g} \\
    \textstyle
    &g_r(\mathbf{W}_e, \mathbf{W}_s) = {\textstyle\sum_{i\in\{e,s\}} \lVert \mathbf{W}_i \rVert_F^2}, \label{eq:g_r}
\end{align}
where \(g_e\) and \(g_s\) enforce the alignment of edge points and sewing-line points, respectively, while \(g_r\) promotes smoothness and numerical stability. The scalar \(\alpha\) is the regularization parameter that controls the strength of \(g_r\) and sets the trade-off between alignment accuracy and the smoothness of \(\mathbf{W}_e\) and \(\mathbf{W}_s\). A smaller \(\alpha\) allows more aggressive weighting to reduce alignment error, whereas a larger \(\alpha\) penalizes large weight variations and yields smoother, more stable \(\mathbf{W}_e\) and \(\mathbf{W}_s\), though it may reduce alignment accuracy.
%
Finally, the optimization is subject to the following constraints
\begin{align}
	\label{eq:const_norm}
	&\textstyle
    \sum_{i \in \{e, s\}} \mathbf{1}_{n_i}^{\top} \mathbf{W}_i^{\top} \mathbf{1}_{n_i} = 1, \\
	\label{eq:const_balancing}
	&\textstyle\sum_{i \in \{e, s\}} 
    (-1)^{\delta_{i,e}}\left\| \mathbf{E}_i \right\|_F^2
    \, \mathbf{1}_{n_{j}}^{\top} \mathbf{W}_{j}^{\top} \mathbf{1}_{n_{j}}
    \,
    = 0, \\
	\label{eq:const_ortho}
	&\mathbf{R}^\top \mathbf{R} = \mathbf{I}_d, \quad \det(\mathbf{R}) = 1, \quad \alpha > 0,\\
	\label{eq:const_bound_weights}
	&\mathbf{W}_i \geq \mathbf{0}, \quad i\in\{e,s\},
\end{align}
where \(j \neq i\) is the complementary index in \(\{e, s\}\), and \(\delta_{i,e}\) is the \textit{Kronecker Delta} function which equals 1 if \(i = e\) and 0 otherwise.
{
\color{black}
Each constraint serves a specific purpose: 
\begin{itemize}
\item \textit{Normalization}~\eqref{eq:const_norm} distributes a fixed total importance among the global and local alignment terms; 
\item \textit{Balancing}~\eqref{eq:const_balancing} prevents either term from dominating by equalizing their weighted contributions; 
\item \textit{Orthonormality}~\eqref{eq:const_ortho} ensures that \(\mathbf{R}\) remains a valid rotation matrix; and
\item \textit{Bounds}~\eqref{eq:const_bound_weights} constrain all weights to be non-negative.
\end{itemize}
}
{
\color{black}
\begin{remark}
    Collectively,~\eqref{eq:min_g}--\eqref{eq:const_bound_weights} define an iterative pose estimation process in which the system continuously refines the fabric panel pose until its edge and sewing line points align accurately with the CAD model.
\end{remark}
}

\subsection{GLW-ICP solution}
The GLW-ICP optimization problem can be solved by dividing it into four steps in each iteration. The problem is structured as a joint optimization over multiple variables—namely, the correspondences \(\mathcal{C}\), the rotation matrix \(\mathbf{R}\), the translation vector \(\mathbf{t}\), and the weight vectors \(\mathbf{w}_e\) and \(\mathbf{w}_s\). In each step, one set of variables is optimized while the others are held fixed.

\subsubsection{Update correspondence mapping}
The first step establishes correspondence mapping \(\mathcal{C}\) between the source points (CAD model global edge and local sewing points) and target points (global edge and local sewing target points). This identifies the nearest corresponding point (red and green dots, Fig.~\ref{fig:icp_formulation}) for each global edge and local sewing point (red and green stars).
At the \(k^{\text{th}}\) iteration, the source points are transformed using rotation matrix \(\mathbf{R}_{k-1}\) and translation vector \(\mathbf{t}_{k-1}\) from the \((k-1)^{\text{th}}\) iteration. The \textit{k-d tree algorithm} ~\cite{924423}, a data structure that efficiently organizes points in a k-dimensional space to quickly find the nearest neighbors, is then used to find the nearest neighbor points. Specifically, the correspondence \(c_l\in\mathcal{C}\) for each point is computed as
\begin{align}
    \textstyle
    &c_l = \arg\min_{c_l} \left\| \mathbf{R} \mathbf{p}_l + \mathbf{t} - \mathbf{q}_{c_l} \right\|^2, \quad\text{where}
    \label{eq:GLWICP_closest_point_compact} \\
    &\mathbf{p}_l = \mathbf{p}_{e,l},\quad \mathbf{q}_{c_l} = \mathbf{q}_{e,c_l}, \quad \text{if } l \in \{1, \dots, n_e\}, \notag \\
    &\mathbf{p}_l = \mathbf{p}_{s,l-n_e},\quad \mathbf{q}_{c_l} = \mathbf{q}_{s,c_l}, \quad \text{if } l \in \{n_e+1, \dots, n_e+n_s\}. \notag
\end{align}

\subsubsection{Update translation vector}
\label{sec:update_translation_vector}
The second step in the GLW-ICP method is to update the translation vector \(\mathbf{t}\). 
By fixing \(\mathcal{C}\), \(\mathbf{R}\), \(\mathbf{w}_e\), and \(\mathbf{w}_s\), and noting that the regularization term is \(\mathbf{t}\), the objective function \(g(\cdot)\) in \eqref{eq:min_g} {therefore, simplifies to the first two terms}, given by \eqref{eq:E_e} and \eqref{eq:E_s}.
Hence, the complete \(\mathbf{t}\)-dependent terms in the expansion of \eqref{eq:E_e} and \eqref{eq:E_s} with \eqref{eq:E_i_frob_norm}
can be jointly written as
\begin{align}
    \label{eq:g_i_t1}
    \textstyle
    &g_{i,\mathbf{t}}(\mathbf{t}) = \operatorname{tr}\left(
        \mathbf{R} \mathbf{P}_i \mathbf{W}_i (\mathbf{1}_{n_i} \mathbf{t}^\top)\right. 
     + \mathbf{t} \mathbf{1}_{n_i}^\top \mathbf{W}_i \mathbf{P}_i^\top \mathbf{R}^\top
     \\
        &+ \left.\mathbf{t} (\mathbf{1}_{n_i}^\top \mathbf{W}_i \mathbf{1}_{n_i}) \mathbf{t}^\top 
    - \mathbf{t} \mathbf{1}_{n_i}^\top \mathbf{W}_i \mathbf{Q}_{i,c}^\top
        - \mathbf{Q}_{i,c} \mathbf{W}_i (\mathbf{1}_{n_i} \mathbf{t}^\top)\right), \notag
\end{align}
where \(i\in\{e,s\}\), and \(g_{i,\mathbf{t}}(\mathbf{t})\) denotes only \(\mathbf{t}\) dependent terms of \eqref{eq:E_e} and \eqref{eq:E_s}. Using the \textit{symmetry property} of terms 1 and 2, and 4 and 5 of \eqref{eq:g_i_t1}, and the \textit{cyclic property of the trace of a matrix}, Eq. \eqref{eq:g_i_t1} becomes
\begin{equation}
\label{eq:g_e_t2}
	\begin{aligned}
	g_{i,\mathbf{t}}=2\,\mathbf{t}^\top(\mathbf{R}\mathbf{P}_i - \mathbf{Q}_{i,c})\mathbf{W}_i\,\mathbf{1}_{n_i}
		+ (\mathbf{1}_{n_i}^\top \mathbf{W}_i \mathbf{1}_{n_i})\,\|\mathbf{t}\|^2.
	\end{aligned}
\end{equation}
By using \eqref{eq:g_e_t2}, the partial derivative of the objective function \(g(\cdot)\) with respect to \(\mathbf{t}\) can be written as
{\small
\begin{equation}
 \textstyle
    \label{eq:g_t_partial}
    \frac{\partial g}{\partial \mathbf{t}} 
    = 2 \sum_{i \in \{e, s\}} \Big( 
        \big(\mathbf{R}\mathbf{P}_i - \mathbf{Q}_{i,c} \big) \mathbf{W}_i \mathbf{1}_{n_i} 
        + \big( \mathbf{1}_{n_i}^{\top} \mathbf{W}_i \mathbf{1}_{n_i} \big) \mathbf{t} 
    \Big).
\end{equation}
}Setting \(\frac{\partial g}{\partial \mathbf{t}} = 0\) and solving for \(\mathbf{t}\) yields
\begin{equation}
    \label{eq:t_final}
    \mathbf{t} = \textstyle\frac{\textstyle\sum_{i \in \{e, s\}} \left( \mathbf{Q}_{i,c} - \mathbf{R} \mathbf{P}_i \right) \mathbf{W}_i \mathbf{1}_{n_i}}{\textstyle\sum_{i \in \{e, s\}} \mathbf{1}_{n_i}^{\top} \mathbf{W}_i \mathbf{1}_{n_i}}.
\end{equation}
{
\color{black}
\begin{remark}
Unlike vanilla ICP, which weights all point pairs equally, GLW-ICP updates the translation $\mathbf{t}$ using two adaptive weight matrices, $\mathbf{W}_e$ and $\mathbf{W}_s$, which separately adjust the influence of global edge and local sewing point pairs that suppress occluded and unreliable points.
\end{remark}
}

\subsubsection{Update rotation matrix}

The third step in the GLW-ICP method is to update the rotation matrix \(\mathbf{R}\).  Since the translation vector \(\mathbf{t}\) has already been updated, the objective function \(g(\cdot)\) can be re-written in terms of the centered points \(\tilde{\mathbf{P}}_{e}\in\mathbb{R}^{3\times n_e}\), \(\tilde{\mathbf{Q}}_{e,c}\in\mathbb{R}^{3\times n_e}\), \(\tilde{\mathbf{P}}_{s}\in\mathbb{R}^{3\times n_s}\), and \(\tilde{\mathbf{Q}}_{s,c}\in\mathbb{R}^{3\times n_s}\), which are obtained by subtracting the centroids of the source and target points. The centered points are defined as
\begin{align}
	\label{eq:P_tilde_e}
    \textstyle
	\tilde{\mathbf{P}}_e &= \mathbf{P}_e - \mathbf{c}_p\mathbf{1}_{n_e}^\top, \quad 
	\tilde{\mathbf{Q}}_{e,c} = \mathbf{Q}_{e,c} - \mathbf{c}_q\mathbf{1}_{n_e}^\top, \notag \\
	\tilde{\mathbf{P}}_s &= \mathbf{P}_s - \mathbf{c}_p\mathbf{1}_{n_s}^\top, \quad 
	\tilde{\mathbf{Q}}_{s,c} = \mathbf{Q}_{s,c} - \mathbf{c}_q\mathbf{1}_{n_s}^\top,
\end{align}
where \(\mathbf{c}_p\in\mathbb{R}^{3}\) and \(\mathbf{c}_q\in\mathbb{R}^{3}\) are weighted centroid of the source and target points, given by
\begin{align}
	\label{eq:c_p}
	\mathbf{c}_p = \frac{\mathbf{P}_e \mathbf{w}_e + \mathbf{P}_s \mathbf{w}_s}
	{\mathbf{1}_{n_e}^\top \mathbf{w}_e + \mathbf{1}_{n_s}^\top \mathbf{w}_s}, \notag 
	\mathbf{c}_q = \frac{\mathbf{Q}_{e,c} \mathbf{w}_e + \mathbf{Q}_{s,c} \mathbf{w}_s}
	{\mathbf{1}_{n_e}^\top \mathbf{w}_e + \mathbf{1}_{n_s}^\top \mathbf{w}_s}.
\end{align}
The centered point matrices \(\tilde{\mathbf{P}}_e\), \(\tilde{\mathbf{Q}}_{e,c}\), \(\tilde{\mathbf{P}}_s\), and \(\tilde{\mathbf{Q}}_{s,c}\) are then used in the subsequent steps of the algorithm to compute the optimal rotation matrix \(\mathbf{R}\). Obtaining \(\mathbf{P}_e\) and \(\mathbf{Q}_{e,c}\) from \eqref{eq:P_tilde_e}, and substituting  them in \eqref{eq:E_e} yields
\begin{align}
\mathbf{E}_e(c, \mathbf{R}, \mathbf{t}) &= \mathbf{R} \tilde{\mathbf{P}}_e - \tilde{\mathbf{Q}}_{e,c} ,
\end{align}
where \((\mathbf{R} \mathbf{c}_p+ \mathbf{t} - \mathbf{c}_q) \mathbf{1}_n^\top=0\) 
since \(\mathbf{t}\) computed by \eqref{eq:t_final}, is used to align the centroids of the source and target points, and corresponding \( \mathbf{R}\) will be computed to minimize the residual error. Hence, \eqref{eq:E_e} and \eqref{eq:E_s} can be jointly re-written as
{\small
\begin{align}
    \label{eq:g_e_R1}
     \textstyle
    g_i = \operatorname{tr}(\mathbf{R} \tilde{\mathbf{P}}_i \mathbf{W}_i \tilde{\mathbf{P}}_i^\top \mathbf{R}^\top
    - 2\mathbf{R} \tilde{\mathbf{P}}_i \mathbf{W}_i \tilde{\mathbf{Q}}_{i,c}^\top +  \tilde{\mathbf{Q}}_{i,c}\mathbf{W}_i \tilde{\mathbf{Q}}_{i,c}^\top),
\end{align}
}where \(i=\{e,s\}\). Using the cyclic property of trace, \(\mathbf{R}^{T}\mathbf{R}=\mathbf{I}\), and ignoring \(\mathbf{R}\)-independent terms, \eqref{eq:g_e_R1} can be written as
\begin{equation}
	\label{eq:g_e_R2}
	g_{i,\mathbf{R}}(\mathbf{R})=- 2 \operatorname{tr}( \mathbf{R} \tilde{\mathbf{P}}_i \mathbf{W}_i \tilde{\mathbf{Q}}_{i,c}^\top ),\quad i\in\{e,s\},
\end{equation}
where \(g_{i,\mathbf{R}}(\mathbf{R})\) denotes the \(\mathbf{R}\) dependent term of \(g_i{}\).
The combined \(\mathbf{R}\)-dependent \(g_{\mathbf{R}}(\mathbf{R})\) can be written as
\begin{equation}
	\label{eq:g_R_R}
	g_{\mathbf{R}}(\mathbf{R})=g_{e,\mathbf{R}}(\mathbf{R})+g_{s,\mathbf{R}}(\mathbf{R})=- 2 \operatorname{tr}\left( \mathbf{R} \mathbf{H}\right),
\end{equation}
where
\(
	\mathbf{H}=\tilde{\mathbf{P}}_e \mathbf{W}_e \tilde{\mathbf{Q}}_{e,c}^\top +\tilde{\mathbf{P}}_s \mathbf{W}_s \tilde{\mathbf{Q}}_{s,c}^\top. 
\)
Minimizing \eqref{eq:g_R_R} to obtain optimal \(\mathbf{R}\), is same as 
is equivalent to maximizing its negative. Therefore, the objective function \(g(\mathbf{R})\) for determining the optimal \(\mathbf{R}\) can be rewritten as
\begin{align}
	\label{eq:min_tr_RH}
	\begin{split}
		\mathbf{R} = \arg \max_{\mathbf{R}} \operatorname{tr} \big( \mathbf{R} \mathbf{H} \big),
		\quad\text{s.t.} \quad \mathbf{R}^\top \mathbf{R} = \mathbf{I}.
	\end{split}
\end{align}
To solve \eqref{eq:min_tr_RH}, \textit{Singular Value Decomposition (SVD)} can be applied since SVD decomposes \(\mathbf{H}\) into rotations \(\mathbf{U}\) and \(\mathbf{V}\) and scaling \(\boldsymbol{\Sigma}\). The optimal \(\mathbf{R}\) is given by
\begin{equation} \label{eq:R_final}
	 \textstyle
    \mathbf{R} = 
	\begin{cases} 
		\mathbf{V} \mathbf{U}^\top, & \text{if } \det(\mathbf{V} \mathbf{U}^\top) \geq 0, \\ 
		\mathbf{V} \mathbf{D} \mathbf{U}^\top, & \text{if } \det(\mathbf{V} \mathbf{U}^\top) < 0,
	\end{cases}
\end{equation}
where the product in \eqref{eq:R_final} aligns the rotations optimally to maximize the matrix trace in \eqref{eq:min_tr_RH}, ensuring \(\mathbf{R}\) is the desired rotation matrix.
However, if \(\det(\mathbf{\mathbf{V} \mathbf{U}^\top}) < 0\), the resulting \(\mathbf{R}\) represents a reflection instead of a proper rotation. To correct this, \(\mathbf{V}\) is adjusted by flipping the sign of its last column by post-multiplying it with \(\mathbf{D}=\text{diag}(1, 1, \dots, -1)\).

{
\color{black}
\begin{remark}
Unlike vanilla ICP, which computes rotation from uniformly weighted point pairs, GLW-ICP constructs the cross-covariance matrix $\mathbf{H} = \tilde{\mathbf{P}}_e \mathbf{W}_e \tilde{\mathbf{Q}}_{e,c}^\top + \tilde{\mathbf{P}}_s \mathbf{W}_s \tilde{\mathbf{Q}}_{s,c}^\top$, combining global edge and local sewing point pairs with adaptive weights.
\end{remark}
}
\subsubsection{Update weight vector}
\label{sec:update_wt_vector}
The final step in the GLW-ICP method is to determining the optimal weight matrices \(\mathbf{W}_e\) and \(\mathbf{W}_s\) under specific constraints. 
The combined alignment error and regularization term for either global (\(i = e\)) or local (\(i = s\)) alignment, as used in \(g(\cdot)\) in~\eqref{eq:g}, can be expressed as
\begin{align}
	\label{eq:g_w_extracted}
    \textstyle
	g_{i,\mathbf{W}_i}(\mathbf{W}_i) = \textstyle\operatorname{tr}\left( \mathbf{E}_i \mathbf{W}_i \mathbf{E}_i^\top \right)+ \alpha\lVert \mathbf{W}_i \rVert_F^2, \hspace{0.5em} i\in\{e,s\}.
\end{align}
By adding and subtracting \(\frac{\|\mathbf{E}_i^\top \mathbf{E}_i\|_F^2}{4\alpha}\) on the right-hand side of \eqref{eq:g_w_extracted}, and using the symmetry of \(\mathbf{W}_i\) and the cyclic trace identity
\(\operatorname{tr}(\mathbf{W}_i^\top \mathbf{E}_i^\top \mathbf{E}_i)=\operatorname{tr}(\mathbf{E}_i \mathbf{W}_i \mathbf{E}_i^\top)\), together with
\(\operatorname{tr}((\mathbf{E}_i^\top \mathbf{E}_i)^2)=\|\mathbf{E}_i^\top \mathbf{E}_i\|_F^2\), let \(\mathbf{G}_i=\mathbf{E}_i^\top \mathbf{E}_i\) denote the alignment error Gram matrix, for which \(\operatorname{tr}(\mathbf{G}_i)=\|\mathbf{E}_i\|_F^2\). Then \eqref{eq:g_w_extracted} can be rewritten as
\begin{align}
    \label{eq:g_w_extracted2}
    \textstyle
    g_{i,\mathbf{W}_i}(\mathbf{W}_i) = \textstyle\alpha \left\|\mathbf{W}_i + \frac{\mathbf{G}_i}{2\alpha}\right\|_F^2 - \frac{\|\mathbf{G}_i\|_F^2}{4\alpha}, i \in \{e,s\}. 
\end{align}
Since the subtracted term $-\frac{\|\mathbf{G}_i\|_F^2}{4\alpha}$ in~\eqref{eq:g_w_extracted2} is independent of $\mathbf{W}_i$, it does not affect the minimization of \(g_{i,\mathbf{W}_i}(\cdot)\). Combining cases $i \in \{e,s\}$ yields unified objective function
\begin{align}
	\label{eq:lagrangian}
    \textstyle
	g_{\mathbf{W}}(\mathbf{W}_e,\mathbf{W}_s) =
	\textstyle\frac{1}{2} \sum_{i \in \{e, s\}} \left\| \mathbf{W}_i + \frac{\mathbf{G}_i}{2\alpha} \right\|_F^2.
\end{align}

To handle equality constraints in the optimization problem, constraints~\eqref{eq:const_norm} and~\eqref{eq:const_balancing} are incorporated into objective function using Lagrange multipliers $\mu$ and $\nu$, where $\operatorname{tr}(\mathbf{G}_e) = \|\mathbf{E}_e\|_F^2$ and $\operatorname{tr}(\mathbf{G}_s) = \|\mathbf{E}_s\|_F^2$, resulting in the following Lagrangian \( \mathcal{L} \)
{\small
\begin{align}
    &\mathcal{L}(\mathbf{W}_e, \mathbf{W}_s, \mu, \nu) =\ 
    g_{\mathbf{W}}(\mathbf{W}_e,\mathbf{W}_s)
    + \mu \bigg(1 - \sum_{i \in \{e, s\}} \mathbf{1}_{n_i}^{\top} \mathbf{W}_i^\top \mathbf{1}_{n_i} \bigg) \notag \\
    &
    \qquad\qquad\qquad\quad\hspace{0.5em}+ \nu \sum_{i \in \{e, s\}} (-1)^{\delta_{i,e}} \operatorname{tr}(\mathbf{G}_i)\, \mathbf{1}_{n_j}^\top \mathbf{W}_j \mathbf{1}_{n_j}.
    \label{eq:Lagrangian_matrix}
\end{align}
}This approach enables the constrained optimization problem to be solved systematically using the \textit{Karush-Kuhn-Tucker (KKT)} conditions. To obtain  \(\mathbf{W}_e\), taking the partial derivatives of \( \mathcal{L} \) with respect to \(\mathbf{W}_e\) yields
\begin{align}
\label{eq:dL_dW}
\textstyle
\frac{\partial \mathcal{L}}{\partial \mathbf{W}_i}
= \textstyle\mathbf{W}_i + \frac{\mathbf{G}_i}{2\alpha}
- \mu\, \mathbf{1}_{n_i} \mathbf{1}_{n_i}^\top
+ \nu\, (-1)^{\delta_{k,e}} \operatorname{tr}(\mathbf{G}_k)\, \mathbf{1}_{n_i} \mathbf{1}_{n_i}^\top,
\end{align}
where \(k \neq i\) is the complementary index in \(\{e, s\}\).
Equating \eqref{eq:dL_dW} to zero, and solving for \( \mathbf{W}_i\) yields
\begin{equation}
	\label{eq:W_e}
	\mathbf{W}_i = 
-\frac{\mathbf{G}_i}{2\alpha}
+ \left[
\mu
- \nu\, (-1)^{\delta_{k,e}} \operatorname{tr}(\mathbf{G}_k)
\right] \mathbf{1}_{n_i} \mathbf{1}_{n_i}^\top. 
\end{equation}
Since the diagonal elements of \(\mathbf{W}_i\) are of interest, the off-diagonal contributions can be ignored, and hence \eqref{eq:W_e} can be simplified as 
\begin{equation}
	\label{eq:W_e2}
	\mathbf{W}_i = -\frac{1}{2\alpha} \operatorname{diag}(\mathbf{G}_i)
    + \left(\mu - \nu\, (-1)^{\delta_{k,e}} \operatorname{tr}(\mathbf{G}_k)\right) \mathbf{I}_{n_i},
\end{equation}
where \(i\in\{e,s\}\), and \(\mathbf{I}_{n_i}\in\mathbb{R}^{n_i\times n_i}\) denotes identity matrix. 

To obtain \(\mu\) and \(\nu\), Eq. \eqref{eq:W_e2} for \(i=e\) and \(i=s\) is substituted into ~\eqref{eq:const_norm} and \eqref{eq:const_balancing} to satisfy the normalization and balancing constraints, respectively, which can be solved to yield
\begin{align}
	\label{eq:mu_2}
	\mu &= \frac{\left( 2\alpha + \beta \right)\left(n_e \operatorname{tr}(\mathbf{G}_s)^2 + n_s \operatorname{tr}(\mathbf{G}_e)^2\right)}{2\alpha\beta^2 n_e n_s},  \\
	\label{eq:nu_2}
	\nu &= \frac{\left( 2\alpha + \beta \right)\left(n_e \operatorname{tr}(\mathbf{G}_s) - n_s \operatorname{tr}(\mathbf{G}_e)\right)}{2\alpha\beta^2 n_e n_s},
\end{align}
where
\(\beta = \operatorname{tr}(\mathbf{G}_e) + \operatorname{tr}(\mathbf{G}_s)\).
Substituting \eqref{eq:mu_2}, and \eqref{eq:nu_2} in \eqref{eq:W_e2} gives
{\small
\begin{align}
	\label{eq:W_e_1}
	\mathbf{W}_i &= 
	\frac{\left( 2\alpha + \beta \right) \operatorname{tr}(\mathbf{G}_i) \mathbf{I}_{n_i} - n_i\,\beta\, \operatorname{diag}(\mathbf{G}_i) }{2\alpha\beta n_i }, i\in\{e,s\}. 
\end{align}
}These weights satisfy both the normalization constraint~\eqref{eq:const_norm} and the global error complementarity constraint~\eqref{eq:const_balancing}, and ensure the minimization of the objective function~\eqref{eq:min_g}.

{
\color{black}
\begin{remark}
As compared to vanilla ICP, two separate weight matrices, $\mathbf{W}_e$ for edges and $\mathbf{W}_s$ for sewing lines, are used to control the relative influence of global edge points and local sewing line points, respectively, in the alignment process.
\end{remark}
}
{
\color{black}
\subsection{Regularization in GLW-ICP}
The influence of \(\alpha\) on the diagonal weights follows directly from \eqref{eq:dL_dW} and its closed-form solution \eqref{eq:W_e}. Specifically, \(\alpha\) appears through the factor \(1/(2\alpha)\) multiplying \(\mathbf{G}_i\), so increasing \(\alpha\) decreases the contribution of the alignment error Gram matrix term in \(\mathbf{W}_i\), given by \eqref{eq:W_e2}. Consequently, a smaller \(\alpha\) allows more aggressive weighting to reduce alignment error, whereas a larger \(\alpha\) penalizes large weight variations and yields smoother, more stable \(\mathbf{W}_i\), at the cost of potentially reduced alignment accuracy. 
In GLW-ICP, \(\alpha\) is updated during iterations to balance the alignment term and regularization.
}

\subsection{Occlusion handling with sparsity in GLW-ICP}
\label{sec:sparse_wt_vector}

Roller end effector occlusions in the RGB image can generate global target edge points $\mathbf{Q}_e$ within the occluded region, which have no correspondence with the CAD model edge points $\mathbf{P}_e$. To mitigate the unmatched points influence during pose estimation, sparsity is introduced in $\mathbf{W}_e$.
The sparsity is controlled by the \textit{sparsity level} $k < n_e \in \mathbb{N}$~\cite{guo2022adaptive}, a hyperparameter specifying the number of non-zero diagonal weights.
To implement sparsity in $\mathbf{W}_e$, the diagonal of $\mathbf{G}_e$ can be written as
\begin{align}
\label{eq:diag_G_e}
    \operatorname{diag}(\mathbf{G}_e) = 
\begin{bmatrix}
    \| \mathbf{e}_1 \|_2^2,\, \| \mathbf{e}_2 \|_2^2,\, \dots,\, \| \mathbf{e}_{n_e} \|_2^2
\end{bmatrix}^\top\in\mathbb{R}^{n_e},
\end{align}
where \( \mathbf{G}_e = \mathbf{E}_e^\top \mathbf{E}_e \), and \( \mathbf{e}_j \) denotes the \(j^{\text{th}}\) column of \( \mathbf{E}_e \).
Arranging the diagonal elements of \eqref{eq:diag_G_e} in ascending order, the result is denoted as
\begin{equation}
    \label{eq:G_r_e}
    \mathbf{G}^{r}_{e} = \operatorname{diag}\left(
    \begin{bmatrix}
        \| \mathbf{e}_{i_1} \|_2^2,\, \dots,\, \| \mathbf{e}_{i_{n_e}} \|_2^2
    \end{bmatrix}^\top
    \right)\in\mathbb{R}^{n_e \times n_e},
\end{equation}
where superscript \(r\) denotes rearranging, \( \mathbf{G}^{r}_{e} = \operatorname{diag}\left( \operatorname{sort}\big( \operatorname{diag}(\mathbf{G}_e) \big) \right)\), and the indices \( i_1, \dots, i_{n_e} \) are chosen such that
\(
\| \mathbf{e}_{i_1} \|_2^2 \leq \| \mathbf{e}_{i_2} \|_2^2 \leq \dots \leq \| \mathbf{e}_{i_{n_e}} \|_2^2.
\)
Thus, for any top \( k \) satisfying \( 1 \leq k\leq n_e \), sorted
\eqref{eq:G_r_e} can be given as
\begin{align}
\label{eq:G_r_e_k2}
\textstyle
\mathbf{G}^r_{e,k} &=
\operatorname{diag}\left(
\begin{bmatrix}
	\| \mathbf{e}_{i_1} \|_2^2,  \dots, \| \mathbf{e}_{i_k} \|_2^2
\end{bmatrix}\right),\\
\label{eq:G_r_e_k_dash2}
\mathbf{G}^r_{e,{k^{\prime}}} &=
\operatorname{diag}\left(
\begin{bmatrix}
	\| \mathbf{e}_{i_{k+1}} \|_2^2,  \dots, \| \mathbf{e}_{i_{n_e}} \|_2^2
\end{bmatrix}\right),
\end{align}
where the index $k^\prime = n_e - k$ denotes the number of remaining elements after selecting the top $k$ sorted entries. Here \(\mathbf{G}^r_{e,k}\in\mathbb{R}^{k\times k}\), \(\mathbf{G}^r_{e,{k^{^\prime}}}\in\mathbb{R}^{(n_e-k)\times (n_e-k)}\), and \(\mathbf{G}^r_{e,k},  \mathbf{G}^r_{e,{k^{^\prime}}}\subseteq \mathbf{G}^r_e\).

With the sorted indices, a permutation matrix \(\boldsymbol{\Lambda}\in\mathbb{R}^{n_e\times n_e}\) can be defined as
\begin{equation}
\label{eq:Lambda}
\boldsymbol{\Lambda}[p, q] =
\begin{cases} 
	1 & \text{if } q = i_p, \\
	0 & \text{otherwise}.
\end{cases}
\end{equation}
where \(p\) and \(q\) denote row and column indices of \(\boldsymbol{\Lambda}\), respectively. For example, if \( i_1 = 3 \), then \(p=1\) and \(q=3\), which gives \(\boldsymbol{\Lambda}(1,3)=1\). 
Using $\boldsymbol{\Lambda}$, a permuted matrix $\mathbf{W}^{r}_{e}\in \mathbb{R}^{n_e\times n_e}$, obtained from \(\mathbf{W}_{e}\), can be partitioned as
\begin{equation}
\label{eq:W_r_e}
    \mathbf{W}^{r}_{e} = \mathbf{\boldsymbol{\Lambda}} \mathbf{W}_e \mathbf{\boldsymbol{\Lambda}}^\top = \begin{bmatrix}
    	\mathbf{W}^{r}_{e,k} & \mathbf{0} \\
    	\mathbf{0} & \mathbf{W}^{r}_{e,k^{\prime}}
    \end{bmatrix},
\end{equation}
where $k$ partitions $\mathbf{W}^{r}_{e}$ into $\mathbf{W}^{r}_{e,k} \in \mathbb{R}^{k\times k}$ and $\mathbf{W}^{r}_{e,k^\prime} \in \mathbb{R}^{(n_e-k)\times (n_e-k)}$, specifying the $k$ global edge points $\mathbf{P}_e$ used in each GLW-\ac{ICP} iteration. $\mathbf{W}^{r}_{e,k^\prime}$ contains weights for the remaining $n_e-k$ entries (indices $\{k+1, \ldots, n_e\}$ in the sorted list).
Hence, for \(i=e\), Eq. ~\eqref{eq:W_e_1} can be re-written as
\begin{align}
    \label{eq:W_r_e_k2}
    \textstyle
    \mathbf{W}^{r}_{e,l} &=
    \frac{
        \left( 2\alpha + \beta_{l} \right) \operatorname{tr}(\mathbf{G}^{r}_{e,l}) \mathbf{I}_{l}
        - l\,\beta_{l}\, \mathbf{G}^{r}_{e,l} 
    }{2\alpha\beta_{l}l},l \in\{k, k^\prime\},
\end{align}
where \(\beta_l = \operatorname{tr}(\mathbf{G}^{r}_{e,l}) + \operatorname{tr}(\mathbf{G}_{s})\).

To satisfy the bounds constraint in~\eqref{eq:const_bound_weights}, i.e., enforce positive weights for the top $k$ entries and non-positive weights for the remaining $n_e-k$ (so that $n_e-k$ points from $\mathbf{P}_e$ can be pruned), the bounds on $\alpha$ can be obtained from~\eqref{eq:W_r_e_k2} by solving
\begin{equation}
    \label{eq:W_r_e_l}
    \mathbf{W}^{r}_{e,l}
    =
        \begin{cases}
        > 0, & l = k, \\
        \leq 0, & l = k^\prime.
        \end{cases}
\end{equation}
Since the entries in \eqref{eq:W_r_e} are sorted in decreasing order, ensuring that the $k^{\text{th}}$ entry is strictly positive and the $(k+1)^{\text{th}}$ entry is non-positive automatically guarantees the required conditions for all top $k$ and remaining $n_e-k$ entries. By applying inequalities in \eqref{eq:W_r_e_l} to the $k^{\text{th}}$ and $(k+1)^{\text{th}}$ diagonal entries of $\mathbf{W}^{r}_{e,k}$ and $\mathbf{W}^{r}_{e,k^\prime}$, respectively, the following bounds for regularization parameter $\alpha$ can be directly obtained
\begin{equation}
\label{eq:alpha_bound}
\alpha_{\text{lb}} < \alpha \leq \alpha_{\text{ub}},
\end{equation}
where
\begin{align}	
\label{eq:alpha_lb}
\textstyle
	\alpha_{\text{lb}} &= \frac{k \beta_k \| \mathbf{e}_{i_k} \|_2^2 - \beta_k \operatorname{tr}(\mathbf{G}^{r}_{e,k})}{2 \operatorname{tr}(\mathbf{G}^{r}_{e,k})},\\
\label{eq:alpha_ub}
\textstyle
	\alpha_{\text{ub}} &= \frac{(k+1) \beta_{k+1} \| \mathbf{e}_{i_{k+1}} \|_2^2 - \beta_{{k+1}} \operatorname{tr}(\mathbf{G}^{r}_{e,{k+1}})}{2 \operatorname{tr}(\mathbf{G}^{r}_{e,{k+1}})}.
\end{align}
Now, to strictly ensure  \(\mathbf{W}^{r}_{e,k} > 0\) and \(\mathbf{W}^{r}_{e,k^{\prime}} =0\) as per \eqref{eq:W_r_e_l}, \(\alpha\) in \eqref{eq:W_r_e_k2} and \eqref{eq:W_e_1} (\(i=e\) ) can be set to \(\alpha_{\text{ub}}\), which yields

\begin{align}
\label{eq:W_r_e_l_case}
\textstyle
\mathbf{W}^{r}_{e, l} &=
    \begin{cases}
    \displaystyle \frac{\left( 2\alpha_{\text{ub}} + \beta_l \right) \operatorname{tr}(\mathbf{G}^{r}_{e,l})\,\mathbf{I}_l - l\, \beta_l\, \mathbf{G}^{r}_{e, l}}{2\alpha_{\text{ub}}\beta_l l}, & l=k, \\[2ex]
    0, & {l=k^\prime.}
    \end{cases}\\
    \label{eq:W_r_s_l_case}
    \mathbf{W}_s &= \frac{\left( 2\alpha_{\text{ub}} + \beta_k \right) \operatorname{tr}(\mathbf{G}_s)\,\mathbf{I}_{n_s} - n_s\,\beta_k\, \operatorname{diag}(\mathbf{G}_s)}{2\alpha_{\text{ub}}\,\beta_k\, n_s}
\end{align}

Eqs. \ref{eq:W_r_e_l_case} and \eqref{eq:W_r_s_l_case} ensure that GLW-ICP focuses on the most reliable point pairs, reducing the impact of the points generated within the fabric occluded region that may cause mismatches.

\subsection{Convergence analysis}\label{subsec:convergence analysis}

Let \( g^{(\kappa)} = g(\mathcal{C}^{(\kappa)}, \mathbf{W}_e^{(\kappa)}, \mathbf{W}_s^{(\kappa)}, \mathbf{R}^{(\kappa)}, \mathbf{t}^{(\kappa)}) \) denote the value of the objective function at iteration \(\kappa\). At each iteration, the algorithm updates the variables in four stages by solving a series of subproblems that do not increase the objective function (Sec.~\ref{sec:glwicp_opt_problem_form}--\ref{sec:update_wt_vector}). 

Given the current transformation and weights, the algorithm finds the nearest neighbor correspondences using a \(k\)-d tree~\cite{924423}, minimizing the alignment error for fixed transformations so that \( g(\mathcal{C}^{(\kappa+1)}) \leq g(\mathcal{C}^{(\kappa)}) \). Next, fixing correspondences and weights, the algorithm computes closed-form solutions for the optimal \(\mathbf{t}^{(\kappa+1)}\) and \(\mathbf{R}^{(\kappa+1)}\), ensuring \( g(\mathbf{R}^{(\kappa+1)}, \mathbf{t}^{(\kappa+1)}) \leq g(\mathbf{R}^{(\kappa)}, \mathbf{t}^{(\kappa)}) \). Finally, the weights \(\mathbf{W}_e\) and \(\mathbf{W}_s\) are updated by solving a constrained optimization problem, which also has a closed-form solution, further reducing the weighted alignment error and enforcing sparsity and regularization, so that \( g(\mathbf{W}_e^{(\kappa+1)}, \mathbf{W}_s^{(\kappa+1)}) \leq g(\mathbf{W}_e^{(\kappa)}, \mathbf{W}_s^{(\kappa)}) \). 

Since each step does not increase the objective and \(g\) is always non-negative (as a weighted sum of squared errors), the sequence \(\{g^{(\kappa)}\}\) is monotonically decreasing and bounded below, and thus converges to a finite value:
\begin{align}
\label{eq:g_kappa}
\lim_{\kappa \to \infty} g^{(\kappa)} = g^* \geq 0.
\end{align}
Hence, the algorithm ensures bounded convergence and provides a stable pose estimate after a finite number of iterations. Algorithm~\ref{alg:GLW-ICP} outlines the GLW-ICP algorithm for estimating the optimal rigid transformation \(\mathbf{R}\) and \(\mathbf{t}\) between the source and target points.

\begin{algorithm}[ht]
	\caption{GLW-ICP Algorithm}
	\label{alg:GLW-ICP}
	\begin{algorithmic}[1]
		
		\STATE \textbf{Input:} Global edge points \({\mathbf{P}}_e\),
		Local sewing points \({\mathbf{P}}_s\), \\
		Global edge target points \({\mathbf{Q}}_e\), Local sewing target points \({\mathbf{Q}}_s\), 
		Sparsity level \(k\), 
		Tolerance \(\epsilon\). 
		\STATE \textbf{Output:} Rotation matrix $\mathbf{R}$, Translation vector $\mathbf{t}$. 
		\STATE Initialize $\mathbf{R} = \mathbf{I} $, $\mathbf{t} = 0$,
		$\mathbf{w}_e =\mathbf{w}_s= \frac{1}{n+m}[1, 1, \dots, 1]^\top$.
		\REPEAT
		\STATE Update correspondence \( {C}_\kappa(x) \) using~\eqref{eq:GLWICP_closest_point_compact}. 
        \STATE Update translation vector $\mathbf{t}$ using~\eqref{eq:t_final}. 
		\STATE Update rotation matrix $\mathbf{R}$ using~\eqref{eq:R_final}.
		\STATE Update weight matrices \(\mathbf{W}_e\) and \(\mathbf{W}_s\) using~\eqref{eq:W_e_1}. 
        {\color{black}\STATE Update regularization parameter $\alpha$ using~\eqref{eq:alpha_ub}.}
		\UNTIL convergence: $\left|g^{(\kappa)} - g^{(\kappa-1)}\right| < \epsilon$,
		\RETURN $\mathbf{R}$, $\mathbf{t}$.
		
	\end{algorithmic}
\end{algorithm}
\section{Proposed system workflow} \label{sec:system_operation}

\begin{figure*}[h!]
	\centering
	\includegraphics[width=1\textwidth]{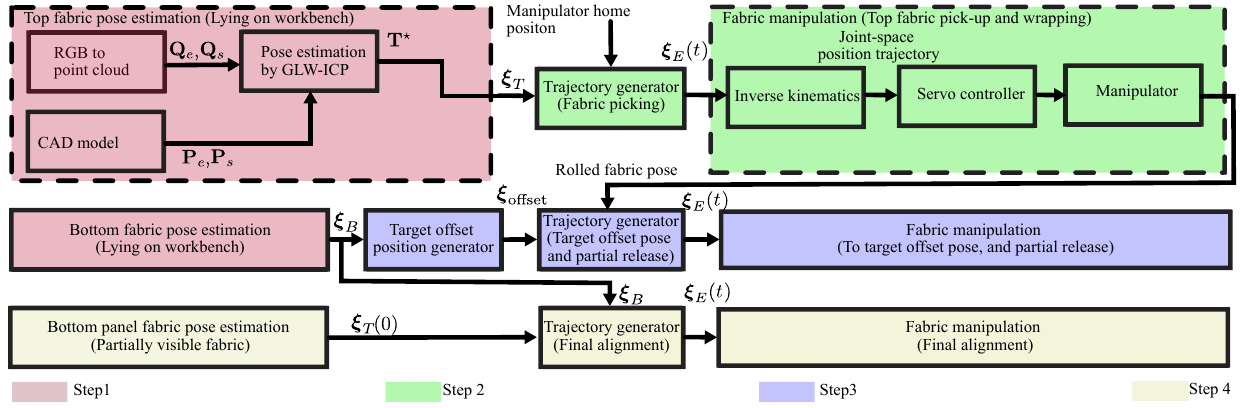} 
	\caption{Complete workflow of the proposed automated fabric alignment system. }
	\label{fig:overall_methodology} 
    \vspace{-3mm}
\end{figure*}

{
\color{black}
In industrial settings, panels are commonly held flat using suction or vacuum tables to maintain wrinkle-minimized surface for repeatable positioning~\cite{kuvzel2022vacuum}. To mimic this configuration and reduce alignment issues, the bottom panel was fixed to the worktable using adhesive tape along the edges. 
}
With this setup, the proposed automated fabric alignment system, as shown in Fig.~\ref{fig:overall_methodology}, proceeds through four main steps.
\subsection{Step 1: Initial pose estimation of bottom and top fabrics}
\label{sec:step_1}

Assuming the top panel fabric is wrinkle free, the camera first captures its image, which is then binarized to extract the fabric edge. The sewing line is subsequently estimated from this edge using an erosion-based image processing method~\cite{soille1999morphological}. The 2D pixel coordinates of the fabric edge and sewing line are mapped to world coordinates using the camera's intrinsic and extrinsic matrices (\({}^W\mathbf{T}_C\))~\cite{zhang2002flexible}, resulting in point clouds \(\mathbf{Q}_e\) and \(\mathbf{Q}_s\) in the world frame \(\mathcal{F}_W\). 

Given the known edge and sewing line point clouds of the CAD model, \(\mathbf{P}_e\) and \(\mathbf{P}_s\), as well as the grasping point predefined relative to the CAD geometry, the GLW-ICP pose estimation method (Sec.~\ref{sec:glwicp_pose_est}) is applied to align the CAD model point cloud with those extracted from both the top and bottom panel fabrics (Fig.~\ref{fig:icp_formulation}). For the top panel, this yields the transformation matrix \({}^W\mathbf{T}_T\) and pose \(\boldsymbol{\xi}_T\); similarly, for the bottom panel, the method provides \({}^W\mathbf{T}_B\) and \(\boldsymbol{\xi}_B\).

\subsection{Step 2: Fabric pick-up and wrapping}
\label{sec:step_2}
Based on the estimated top panel fabric pose $\boldsymbol{\xi}_T$ (Step~1), the target grasping point for the roller-based end-effector is identified on the actual top panel fabric as predefined on the CAD model. 
A fifth-order polynomial trajectory~\cite{11091470}, is used to compute the manipulator joint positions via \ac{IK}, enabling movement from the home position to the target grasping point, following the framework of Kobayashi et al.~\cite{kobayashi2025rollup}. The trajectory is sent to the servo controller, which moves the manipulator and end-effector to the fabric. Upon arrival, the end-effector presses down and activates suction to securely attach to the top panel.

To wrap the fabric onto the roller, the system executes a coordinated motion in which the roller rotates and the end-effector translates synchronously, which allows the fabric to wrap smoothly around the roller without wrinkles, while maintaining consistent pressure against the workbench.

\subsection{Step 3: Wrapped fabric manipulation and partial release}
\label{sec:step_3}
After fully wrapping the fabric, the roller end-effector is moved to an offset pose \({\boldsymbol{\xi}_{\text{offset}}}\), by a similar {trajectory generation discussed in Step 2}, where the orientation is set to match the bottom panel fabric (\( {}^{W}\mathbf{R}_E = {}^{W}\mathbf{R}_B \)), and the position offset is typically set to half the fabric length. At this offset pose, the top panel fabric is partially released.

At this pose, the tilt angles ($\theta_{x}$ and $\theta_{y}$) of the top panel fabric are aligned with those of the bottom panel, while the height $z$ remains the same for both panels since they lie on the same plane. Therefore, the final alignment task reduces to matching the remaining elements of $\boldsymbol{\xi}_T$ with those of $\boldsymbol{\xi}_B$ by dragging the top panel on the bottom panel.
This strategy is adopted to minimize unnecessary end-effector rotations in the subsequent step, so that primarily translational motion is required for final alignment. As a result, the risk of fabric wrinkling during manipulation is significantly reduced.

\subsection{Step 4: Fabric pose estimation and fabric final alignment} 
\label{sec:step_4}
 
After partial fabric release, the state of top fabric panel is re-estimated using GLW-ICP, as its deformable nature prevents it from maintaining its original shape. Additionally, suction applied by the roller further alters the fabric's state, making pose estimation after partial release necessary.
At time-step $t=0$ of the final alignment, $\boldsymbol{\xi}_T(0)$ is obtained in Step~1, except that the point cloud of the partially visible fabric is extracted from RGB image to match the CAD model's edge and sewing line point clouds, as discussed in Sec.~\ref{sec:fabric_pose_est}.

For successful fabric alignment at time-step \(t_f\), the top panel fabric's initial and final poses, given by 
\(\boldsymbol{\xi}_T(0)\) and \(\boldsymbol{\xi}_T(t_f)=\boldsymbol{\xi}_B\),
respectively, are used to generate the fifth-order polynomial trajectory for the end-effector from \( t = 0 \) to \( t = t_f \), \(\boldsymbol{\xi}_E(t)\).
The robot end-effector then follows the computed trajectory to drag the top panel fabric, ensuring smooth, continuous motion to prevent wrinkles. During dragging, the roller maintains suction on the fabric to prevent any relative motion between the top panel and the end-effector.
Once local alignment between the top and bottom panel fabrics is achieved, the system performs a controlled fabric release while simultaneously adjusting the end-effector's position. This step completes the fabric alignment for sewing.
\section{Pose estimation and fabric alignment results}
\label{sec:pose_est_and_fabric}

In this section, the proposed GLW-ICP method's pose estimation accuracy was first evaluated by varying sparsity level \(k\), which prioritizes most accurate point correspondences. 
Specifically, only the $k$ pairs of points with the smallest alignment errors (where $k < n_e$) are assigned non-zero weights, while all other pairs have weights set to zero. This ensures that the registration prioritizes the most accurate matches (Sec.~\ref{sec:sparse_wt_vector}).
Then, regularization parameter \(\alpha\) effect on the GLW-ICP was analyzed.
Next, GLW-ICP pose-estimation accuracy was compared with the baseline methods (Sec.~\ref{sec:existing_pose_estimation}).
%
The complementary roles of global edge and local sewing-line
features in GLW-ICP was validated in Sec. \ref{sec:sewing_line_complementarity}.
Finally, real-world fabric alignment experiments were conducted according to the work-flow discussed in Sec. \ref{sec:system_operation}, to evaluate the performance of the proposed fabric alignment system.

\subsection{Dataset description and setup}\label{sec:dataset}
\begin{figure}[t]
    \centering
    \includegraphics[width=1\linewidth]{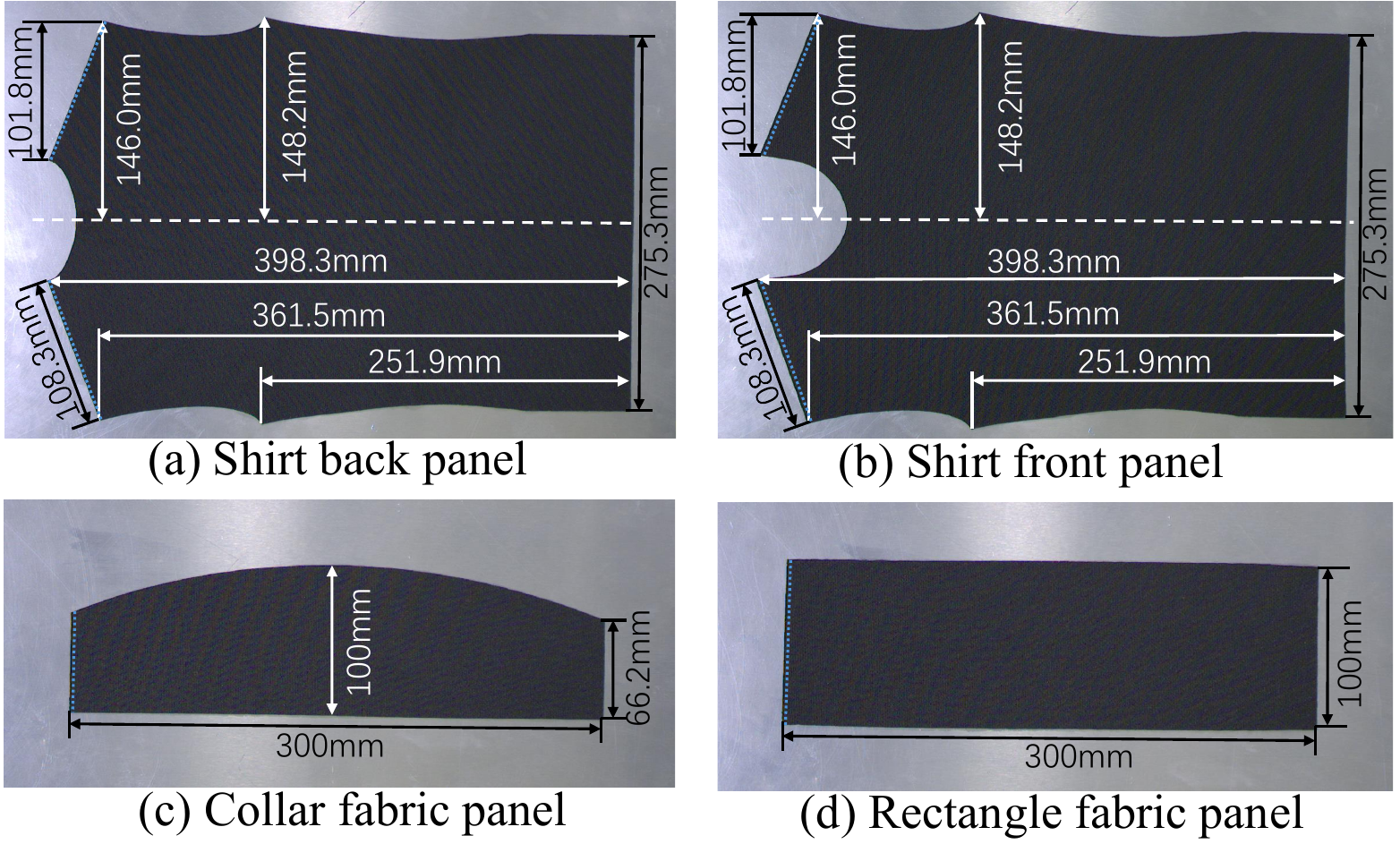}
    \caption{Real fabric samples.}
    \label{fig:fabric_samples}
    \vspace{-3mm}
\end{figure}

For GLW-ICP evaluation, a dataset was constructed from four representative fabric panel shapes common in garment manufacturing: \textit{collar}, \textit{rectangle}, and \textit{shirt}, as shown in Fig.~\ref{fig:fabric_samples}. 
{
\color{black}
These shapes were selected as they represent frequently produced components in industrial garment assembly, covering straight, curved, and mixed edge characteristics. While more complex shapes can be accommodated, the present work focuses on these representative flat components for direct practical relevance to real sewing operations.
} All panels were cut from plain-weave polyester material with the following measured mechanical parameters: fabric thickness \(T = 0.5~\mathrm{mm}\), bending stiffness \(B = 3.09~\mathrm{mN{\cdot}mm}\), and in-plane elastic modulus \(E_{22} = 0.29~\mathrm{MPa}\).{
\color{black}
During experiments, each fabric panel was fixed to a flat stainless-steel using adhesive tape along the edges, mimicking the vacuum holding used in industries and preventing slip and major wrinkles during the experiments.
}

Next, three alignment experiments were defined: shirt alignment (aligning shirt front and back panels with different geometries), collar alignment, and rectangular panel alignment (both using identical panels). In each experiment, the bottom panel is fixed and the top panel is manipulated by the manipulator in the fabric alignment system. 

As discussed in Sec.~\ref{sec:step_1}, the edge and sewing line points of the top panel fabric (\(\mathbf{Q}_e\) and \(\mathbf{Q}_s\)) were extracted and aligned to their corresponding CAD model points (\(\mathbf{P}_e\) and \(\mathbf{P}_s\)) using the proposed GLW-ICP and baseline methods. The resulting optimal rigid transformation \((\mathbf{R}^\star, \mathbf{t}^\star)\) was then used to compute the estimated points \(\hat{\mathbf{Q}}_l = \mathbf{R}^\star \mathbf{P}_l + \mathbf{t}^\star \mathbf{1}_{m_l}^\top\), where \(l \in \{e, s\}\).{
\color{black}
To ensure consistent stopping conditions, all experiments used a convergence tolerance of \(\epsilon = 10^{-7}\) and were repeated for ten trials.
}

\subsection{Error evaluation metrics}
To evaluate \textit{alignment accuracy} after fabric pose estimation, the bi-directional \textit{alignment Chamfer Root Mean Square Error (RMSE)} between estimated points $\hat{\mathbf{Q}}_l$ and target points $\mathbf{Q}_l$ was computed. This metric quantifies the average geometric discrepancy between estimated and ground-truth points, and is defined as
\begin{align}
    \begin{split}
        \label{eq:e_chamfer}
        \textstyle
        e_{\mathrm{Chamfer}}(\hat{\mathbf{Q}}_l, \mathbf{Q}_l) &= 
        \sqrt{ \frac{1}{2} \left( e_{\hat{\mathbf{Q}}_l} + e_{\mathbf{Q}_l} \right) },\quad  l\in \{e, s\},
    \end{split}
\end{align}
where
\(
e_{\hat{\mathbf{Q}}_l} = \frac{1}{m_l} \textstyle\sum_{i=1}^{m_l} \min_{1 \leq j \leq n_l} \left\| {\mathbf{\hat q}}_{l,i} - \mathbf{q}_{l,j} \right\|^2, 
        e_{\mathbf{Q}_l} = \frac{1}{n_l} \textstyle\sum_{j=1}^{n_l} \min_{1 \leq i \leq m_l} \left\| \mathbf{q}_{l,j} - \hat{\mathbf{q}}_{l,i} \right\|^2
\).
Here \(e_{\hat{\mathbf{Q}}_l}\) and \(e_{\mathbf{Q}_l}\) measure alignment error as the mean squared distance to the closest point in the other set.

\subsection{Effect of sparsity level and optimal \texorpdfstring{$k_{\text{ratio}}$}{k-ratio} selection}

To evaluate the effect of \(k\) (Sec.~\ref{sec:sparse_wt_vector}) on removing edge points within the occluded top fabric region, those without correspondence to CAD model edge points, the \textit{sparsity level ratio} \(k_{\text{ratio}} = \frac{k}{n_{e}}\) was used to determine the proportion of top fabric edge points \(\mathbf{Q}_e\) retained.
Hence, sewing line alignment accuracy, measured by alignment Chamfer RMSE \(e_\mathrm{Chamfer}(\hat{\mathbf{Q}}_s, \mathbf{Q}_s)\), was evaluated as a function of the sparsity level \(k_{\text{ratio}}\). For comparison across varying fabric types and alignment scenarios, alignment Chamfer RMSE values were normalized using min-max scaling, referred to as \textit{normalized alignment Chamfer RMSE}. This was computed by subtracting $\min(e_{\mathrm{Chamfer}})$ from $e_{\mathrm{Chamfer}}$, then dividing by $\max(e_{\mathrm{Chamfer}}) - \min(e_{\mathrm{Chamfer}})$.

\begin{figure}[!t]
    \centering
    \includegraphics[width=1\linewidth]{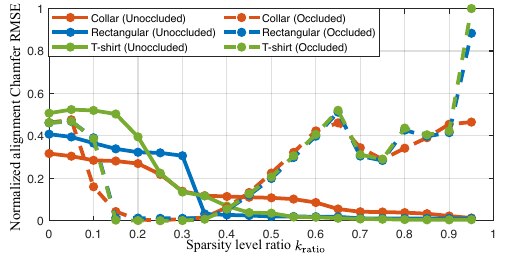}
    \caption{Sparsity level \(k_{\text{ratio}}\) effect on pose estimation accuracy.}
    \label{fig:k_ratio}
    \vspace{-3mm}
\end{figure}

\begin{figure}[!h]
    \centering
    \includegraphics[width=1\linewidth]{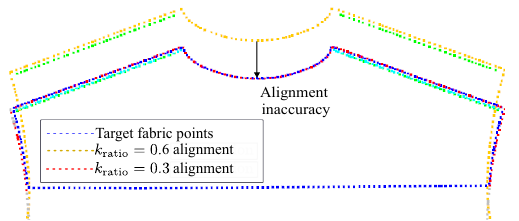}
    \caption{Effect of \(k_{\text{ratio}} \) on pose estimation results.}
    \label{fig:k_ratio_effect}
    \vspace{-3mm}
\end{figure}

Experiments were conducted for \(k_{\text{ratio}} \in [0, 0.95]\) in steps of 0.05, using three fabric shapes (Sec.~\ref{sec:dataset}) under two scenarios: \emph{Unoccluded}, where the entire fabric contour is visible, and \emph{Occluded}, where parts of the fabric were partially hidden by end-effector interaction.
%
%
Under the unoccluded condition, increasing \(k_{\text{ratio}}\) decreases \(e_\mathrm{Chamfer}\) and improves alignment accuracy for all fabric shapes (Fig.~\ref{fig:k_ratio}). This improvement occurs because a higher \(k_{\text{ratio}}\) retains more valid edge points, providing better geometric information for accurate alignment.
However, under the occluded condition, the trend is non-monotonic. As \(k_{\text{ratio}}\) increases from 0 to around 0.30, alignment accuracy improves for all fabric shapes (Fig.~\ref{fig:k_ratio_effect}, red dotted line). This is because the increase in \(k_{\text{ratio}}\) allows the algorithm to retain more edge points from visible regions, thereby providing richer and more complete geometric information for pose estimation. 
Beyond \(k_{\text{ratio}} = 0.35\), increasing \(k_{\text{ratio}}\) reintroduces edge points from occluded areas, which have no correspondences with target points, causing false matches and higher \(e_\mathrm{Chamfer}\) (Fig.~\ref{fig:k_ratio_effect}, yellow dotted line).
Based on these results, the same sparsity level ratio was used for all fabric shapes, with \(k_{\text{ratio}}=0.95\) for unoccluded cases to maximize alignment accuracy and \(k_{\text{ratio}}=0.35\) for occluded cases to retain reliable correspondences.
\begin{remark}
The optimal sparsity level ratio \(k_{\mathrm{ratio}}\) was determined experimentally under each visibility condition (unoccluded and occluded) and was found to be consistent across all fabric shapes (collar, rectangle, and shirt). Once established, this value was fixed for all repeated trials.
\end{remark}
{\color{black}
            Because $k_{\mathrm{ratio}} = k/n_e$ is a ratio of visible edge points to the total edge points, its value is determined by the visible fraction of the edge rather than by the fabric material. If the end-effector or release configuration changes, $k_{\mathrm{ratio}}$ should be re-calibrated by repeating the procedure.
}

\subsection{Effect of regularization parameter \(\alpha\)}
\label{sec:alpha_effect}

\begin{figure}[!t]
    \centering
    \includegraphics[width=1\linewidth]{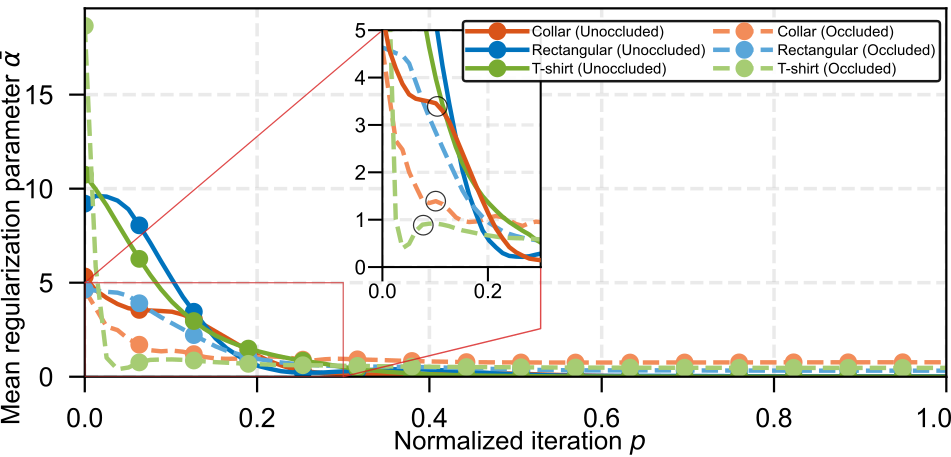}
    \caption{Mean regularization parameter \(\bar{\alpha}(p)\) over iteration.}
    \label{fig:alpha}
\end{figure}

In GLW-ICP, \(\alpha\) was set to the upper bound \(\alpha_{\mathrm{ub}}\), given by \eqref{eq:alpha_ub} to enforce positive weights for top \(k\) correspondences while suppressing the remaining matches.
To characterize the evolution of $\alpha$ during optimization, experiments were performed on three fabric shapes under both unoccluded and occluded visibility conditions. In Fig.~\ref{fig:alpha}, the iteration count was normalized to $p\in[0,1]$, denoted as \textit{normalized iteration}. The x-axis shows $p$, and the y-axis reports the \textit{mean regularization parameter} $\bar{\alpha}(p)$, averaged over ten trials for each fabric shape and visibility condition. $\bar{\alpha}(p)$ was computed as
\begin{equation}
    \bar{\alpha}(p) = \frac{1}{N} \sum_{i=1}^{N} \alpha_i(p),  \quad p\in[0,1],
    \label{eq:alpha_mean}
\end{equation}
where \(\alpha_i(p)\) is the value of \(\alpha\) at iteration \(p\) in the \(i\)-th trial.

As shown in Fig.~\ref{fig:alpha}, $\bar{\alpha}$ increases in the early iterations (circled) and then stabilizes, where the circled rise reflects a temporary regularization adjustment for two reasons: (i) as new correspondences form along curved edges, the error distribution changes, increasing the residuals $\|\mathbf{e}_{i_k}\|_2^2$ in \eqref{eq:alpha_ub} and thus raising $\alpha_{\text{ub}}$; and (ii) the regularizer briefly strengthens the smoothness term to damp weight oscillations during these rapidly changing matches.
As correspondences stabilize and residuals drop, $\bar{\alpha}$ converges to a steady value.

\subsection{Pose estimation comparison with baseline methods}
%
In this section, the GLW-ICP pose-estimation accuracy is compared with the baseline ICP methods (Table \ref{tab:comparison_summary_combined}), briefly discussed in Sec.~\ref{sec:existing_pose_estimation}.
Using these methods, the pose estimation accuracy was evaluated using the alignment Chamfer RMSE between aligned contours and ground-truth, separately for edge and sewing line points under both unoccluded and partially occluded conditions, using~\eqref{eq:e_chamfer}.

\paragraph*{Quantitative results} 

\color{black}
The shaded columns of Table~\ref{tab:comparison_summary_combined} summarizes the pose estimation results across all fabric shapes under both visibility conditions (unoccluded and occluded) for GLW-ICP and all baseline methods. From Table~\ref{tab:comparison_summary_combined}, GLW-ICP succeeds in all cases, including under occlusion. Among the baselines, AW-RICP is the only method that also succeeds consistently under occlusion; however, under occlusion GLW-ICP reduces the Chamfer error by at least \(\sim\)50\% relative to AW-RICP.
Notably, several methods such as Sparse-ICP, KSS-ICP, CPD and FRICP exhibit flipping issues even in the unoccluded setting, leading to misaligned orientations, highlighted in Table \ref{tab:comparison_summary_combined}.
Overall, GLW-ICP achieves the strongest pose estimation accuracy across fabric shapes, consistently producing the lowest Chamfer errors. 

{
\color{black}
Additionally, Table~\ref{tab:comparison_summary_combined} shows that GLW-ICP converges in a moderate number of iterations (typically 30--45) with a total runtime of about 1--3\,s. It is generally comparable in efficiency to other methods, though not consistently the fastest since simpler ICP style approaches can converge in fewer iterations or shorter runtime, reflecting a trade-off between accuracy and speed.
}
{
\color{black}
Finally, paired $t$-tests confirmed that the improvements achieved by GLW-ICP over the baseline methods were statistically significant under all visibility conditions ($p < 0.01$).
}
\newcommand{\msS}{\textcolor{green!60!black}{\textbf{S}}}
\newcommand{\msF}{\textcolor{red!70!black}{\textbf{F}}}
\definecolor{hlrect}{RGB}{255,243,205} 

\newcommand{\hlcell}[1]{\cellcolor{hlrect}{#1}}

\begin{table*}[t]
\centering
\setlength{\abovecaptionskip}{6pt}
\setlength{\belowcaptionskip}{6pt}
\caption{Pose estimation and final alignment comparison across fabric shapes.}
\label{tab:comparison_summary_combined}
\resizebox{\textwidth}{!}{%
\begin{tabular}{
ll
*{9}{>{\columncolor{gray!10}}c}  
c
}
\hline
\multirow{3}{*}{\textbf{Fabric}} &
\multirow{3}{*}{\textbf{Method}} &
\multicolumn{9}{c}{\cellcolor{gray!10}\textbf{Pose estimation}} &
\multirow{2}{*}{\textbf{Final alignment}} \\
\cline{3-11}
& &
\cellcolor{gray!10}{\(e_{\mathrm{Chamfer}}\)} (mm) &
\cellcolor{gray!10}{\(e_{\mathrm{Chamfer}}\)} (mm) &
\cellcolor{gray!10}{Success/Failure} &
\cellcolor{gray!10}{\(e_{\mathrm{Chamfer}}\)} (mm) &
\cellcolor{gray!10}{Success/Failure} &
\cellcolor{gray!10}{Runtime (s)} &
\cellcolor{gray!10}{Iterations}  &
\cellcolor{gray!10}{Runtime (s)} &
\cellcolor{gray!10}{Iterations}  &
\textbf{(mm)} \\
\cline{3-11}
& &
\cellcolor{gray!10}\small (Edge, U) &
\cellcolor{gray!10}\small (Sew, U) &
\cellcolor{gray!10}\small (S/F) &
\cellcolor{gray!10}\small (Sew, O) &
\cellcolor{gray!10}\small (S/F) &
\cellcolor{gray!10}\small (U) &
\cellcolor{gray!10}\small (U) &
\cellcolor{gray!10}\small (O) &
\cellcolor{gray!10}\small (O) &
\\
\hline

\multirow{8}{*}{Collar}
& ICP~\cite{besl1992method}              & 4.945$\pm$0.345 & 3.895$\pm$0.329 & \msF & 111.283$\pm$14.210 & \msF & \textbf{0.14$\pm$0.03} & 48$\pm$3.6 & \textbf{0.05$\pm$0.02} & \textbf{20.8$\pm$7.1} & --- \\
& Trimmed-ICP~\cite{chetverikov2002trimmed} & 8.820$\pm$0.201 & 20.900$\pm$0.630 & \msF & 171.502$\pm$42.230 & \msF & 0.40$\pm$0.01 & 117.7$\pm$2.1 & 0.08$\pm$0.03 & 28.4$\pm$11.5 & --- \\
& Sparse-ICP~\cite{bouaziz2013sparse}       & 3.931$\pm$0.351 & 7.762$\pm$0.739 & \msF & 124.767$\pm$24.861 & \msF & 119$\pm$6.12 & 492.5$\pm$11.4 & 0.24$\pm$0.03 & 87.2$\pm$6.8 & --- \\
& KSS-ICP~\cite{10061449}          &\hlcell{ \textbf{1.111$\pm$0.081}} &\hlcell{216.187$\pm$14.275} & \hlcell{\msF} & 107.693$\pm$1.016 &\msF & 11.39$\pm$0.32 & \textbf{15.6$\pm$1.3} & 11.38$\pm$0.04 & 25.5$\pm$1.0 & --- \\
& CPD~\cite{myronenko2010point}              &\hlcell{ 4.083$\pm$0.044} & \hlcell{297.451$\pm$0.067} & \hlcell{\msF} & 56.267$\pm$1.163 & \msF & 22.90$\pm$3.88 & 50.5$\pm$8.3 & 73.43$\pm$3.16 & 397.0$\pm$13.2 & --- \\
& FRICP~\cite{9336308}            & 3.721$\pm$0.342 & 6.728$\pm$0.646 & \msF & 133.607$\pm$16.801 & \msF & 1.51$\pm$0.04 & 29.8$\pm$0.6 & 4.085$\pm$1.81 & 68.9$\pm$5.3 & --- \\
& AW-RICP~\cite{guo2022adaptive}          & 1.239$\pm$0.016 & 2.012$\pm$0.018 & \msS & 3.061$\pm$0.015 & \msS & 6.62$\pm$0.38 & 168.9$\pm$11.0 & 2.54$\pm$0.84 & 63.1$\pm$20.1 & --- \\
& \textbf{GLW-ICP (ours)}          & 1.202$\pm$0.027 & \textbf{1.974$\pm$0.015} & \msS & \textbf{1.544$\pm$0.012} & \msS & 1.32$\pm$0.29  & 35.9$\pm$1.6 & 2.71$\pm$0.17 & 52.8$\pm$3.9 & \textbf{1.286$\pm$0.527} \\
\hline

\multirow{8}{*}{Rectangle}
& ICP              & 1.124$\pm$0.090 & 1.861$\pm$0.021 & \msS & 131.750$\pm$16.150 & \msF & \textbf{0.12$\pm$0.004} & 38.6$\pm$0.7 & 0.15$\pm$0.04 & 62.6$\pm$19.8 & --- \\
& Trimmed-ICP      & 24.041$\pm$0.280 & 54.411$\pm$0.640 & \msF & 193.450$\pm$13.651 & \msF & 0.28$\pm$0.008 & 79.9$\pm$3.5 & \textbf{0.09$\pm$0.06} & 37.3$\pm$22.2 & --- \\
& Sparse-ICP & \hlcell{1.083$\pm$0.073} & \hlcell{297.252$\pm$0.061} & \hlcell{\msF} & 141.258$\pm$19.324 & \msF & 6.26$\pm$1.17 & 2076.2$\pm$299.6 & 0.36$\pm$0.03 & 142.7$\pm$17.2 & --- \\
& KSS-ICP          & \hlcell{1.027$\pm$0.100} &\hlcell{ 186.148$\pm$142.817} & \hlcell{\msF} & 131.445$\pm$13.110 & \msF & 12.01$\pm$0.48 & 57.8$\pm$4.3 & 11.38$\pm$1.19 & \textbf{12.7$\pm$5.5} & --- \\
& CPD              & \hlcell{3.061$\pm$0.093} & \hlcell{296.323$\pm$0.047} & \hlcell{\msF} & 52.921$\pm$1.830 & \msF & 30.54$\pm$1.08 & 58.9$\pm$2.1 & 60.89$\pm$2.06 & 301.5$\pm$4.3 & --- \\
& FRICP & \hlcell{\textbf{1.024$\pm$0.105}} & \hlcell{149.327$\pm$147.552} & \hlcell{\msF} & 132.060$\pm$10.658 & \msF & 1.70$\pm$0.25 & \textbf{30.5$\pm$4.1} & 4.83$\pm$0.83 & 92.7$\pm$17.1 & --- \\
& AW-RICP          & 1.115$\pm$0.061 & 1.871$\pm$0.083 & \msS & 2.382$\pm$0.114 & \msS & 4.08$\pm$1.34 & 91.1$\pm$13.3 & 2.68$\pm$0.52 & 67.8$\pm$38.7 & --- \\
& \textbf{GLW-ICP (ours)}           & 1.030$\pm$0.091 & \textbf{1.784$\pm$0.044} & \msS & \textbf{1.367$\pm$0.085} & \msS & 1.82$\pm$0.21 & 35.6$\pm$2.4 & 2.97$\pm$0.44 & 58$\pm$5.7 & \textbf{1.311$\pm$0.525} \\
\hline

\multirow{8}{*}{Shirt}
& ICP              & 0.795$\pm$0.054 & 0.812$\pm$0.028 & \msS & 164.850$\pm$13.880 & \msF & \textbf{0.03$\pm$0.002} & 24.9$\pm$0.31 & \textbf{0.05$\pm$0.01} & 45.8$\pm$2.3 & --- \\
& Trimmed-ICP      & 83.868$\pm$67.142 & 127.670$\pm$102.034 & \msF & 76.820$\pm$21.764 & \msF & 0.07$\pm$0.03 & 49.7$\pm$25.7 & 0.08$\pm$0.04 & 60.9$\pm$29.9 & --- \\
& Sparse-ICP       & 0.803$\pm$0.049 & 0.841$\pm$0.063 & \msS & 133.456$\pm$19.123 & \msF & 0.54$\pm$0.03 & 394.9$\pm$24.7 & 0.09$\pm$0.04 & 86.3$\pm$10.7 & --- \\
& KSS-ICP          & \textbf{0.777$\pm$0.048} & \textbf{0.782$\pm$0.065} & \msS & 163.792$\pm$0.584 & \msF & 8.23$\pm$0.51 & \textbf{15.3$\pm$1.1} & 5.934$\pm$0.04 & \textbf{11.1$\pm$1.2} & --- \\
& CPD              & 2.079$\pm$0.102 & 1.911$\pm$0.022 & \msS & 72.999$\pm$0.748 & \msF & 8.21$\pm$0.45 & 49.3$\pm$3.0 & 7.13$\pm$0.07 & 99.9$\pm$0.5 & --- \\
& FRICP            & 165.604$\pm$0.154 & 220.561$\pm$0.414 & \msF & 109.609$\pm$4.536 & \msF & 0.33$\pm$0.03 & 54.7$\pm$5.1 & 0.33$\pm$0.05 & 58.1$\pm$6.2 & --- \\
& AW-RICP          & 0.806$\pm$0.060 & 0.846$\pm$0.086 & \msS & 2.084$\pm$0.028 & \msS & 1.07$\pm$0.07 & 99.4$\pm$17.5 & 0.93$\pm$0.13 & 93.6$\pm$2.1 & --- \\
& \textbf{GLW-ICP (ours)}           & \textbf{0.777$\pm$0.056} & 0.791$\pm$0.078 & \msS & \textbf{0.720$\pm$0.019} & \msS & 0.43$\pm$0.02 & 43.8$\pm$4.0 & 0.35$\pm$0.11 & 25.6$\pm$4.7 & \textbf{1.100$\pm$0.343} \\
\hline
\multicolumn{12}{l}{\textbf{Edge:} Edge points, \textbf{Sew:} Sewing line points. 
\textbf{\textcolor{green!60!black}{S}:} Success, \textbf{\textcolor{red!70!black}{F}:} Failure. 
\textbf{Trials:} 10. \textbf{Note:} {All numeric quantities} are reported as mean $\pm$ standard deviation. 
\textbf{Note:} Best in each column is shown in bold.}
\end{tabular}%
}
\end{table*}

\subsubsection*{Qualitative results}
\begin{figure}[!t]
    \centering
    \includegraphics[width=1\linewidth]{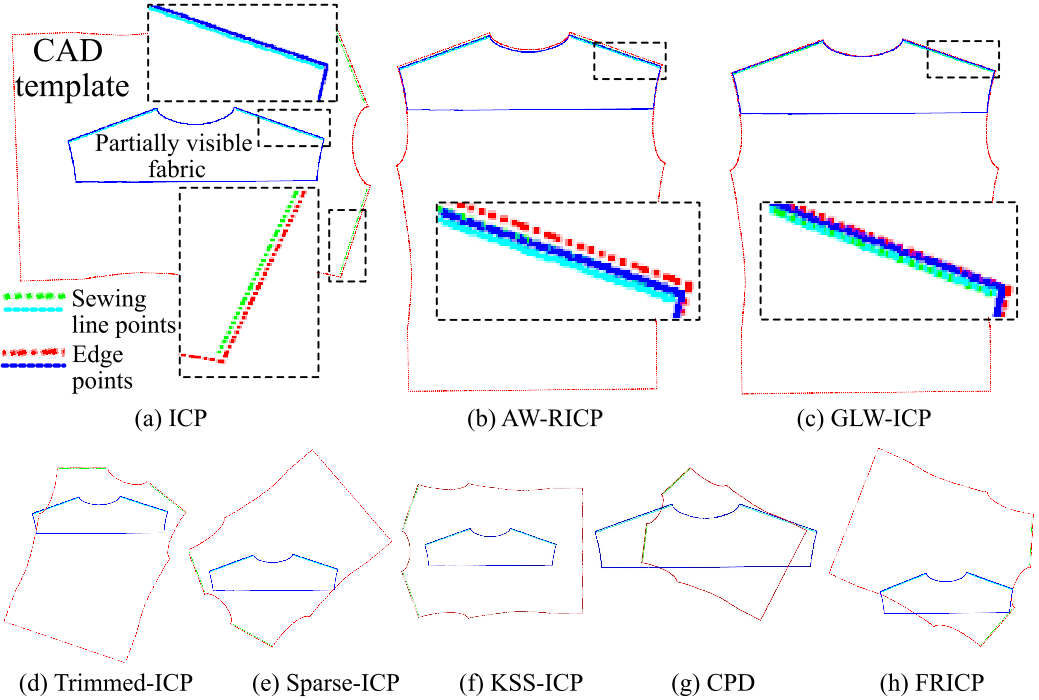}
    \caption{Qualitative pose estimation results.}
    \label{fig:registration_visuals}
    \vspace{-3mm}
\end{figure}

Fig.~\ref{fig:registration_visuals} shows qualitative results for shirt panel pose estimation, where the CAD model’s edge and sewing line points are aligned to those of the partially visible fabric.
%
%
The proposed GLW-ICP demonstrates superior performance across all fabric shapes, consistently outperforming baseline methods in both qualitative and quantitative evaluations. In Fig.~\ref{fig:registration_visuals} (b), GLW-ICP achieves precise alignment under both unoccluded and partially occluded conditions by leveraging global edge features and local sewing lines.
In contrast, most ICP-style baselines fail completely (Fig.~\ref{fig:registration_visuals} (a), (d)--(e)) under occlusion; AW-RICP (Fig.~\ref{fig:registration_visuals} (b)) is the most robust among them, but still does not match the accuracy of GLW-ICP. These observations are further corroborated by Table~\ref{tab:comparison_summary_combined}.

{\color{black}
\subsection{Validation of edge and sewing-line complementarity}
\label{sec:sewing_line_complementarity}

In industrial garment production, tool wear and manual cutting variability often cause local geometric distortions, especially around the shoulder edge. These imperfections make the outer fabric edge deviate from the actual CAD geometry. Using shirt fabric panel, this section validates the complementary roles of global edge and local sewing-line features in the proposed method.

To mimic random cutting error at the shoulder edges, Gaussian noise was added to the shoulder-edge-point subset of the top fabric panel edge points. Let $\mathcal{J}_{\mathrm{shoulder}} \subset \{1,\ldots,n_e\}$ index the shoulder columns of $\mathbf{Q}_e$, so that $\mathbf{Q}_{e,\mathrm{shoulder}}=\mathbf{Q}_e(:,\mathcal{J}_{\mathrm{shoulder}})$. The \textit{perturbed shoulder edge target point matrix} is
\begin{equation}
\tilde{\mathbf{Q}}_{e,\mathrm{shoulder}} = \mathbf{Q}_{e,\mathrm{shoulder}} + \boldsymbol{\zeta},
\quad \boldsymbol{\zeta}\in\mathbb{R}^{3\times |\mathcal{J}_{\mathrm{shoulder}}|},
\label{eq:edge_noise}
\end{equation}
where $\boldsymbol{\zeta}$ is zero-mean Gaussian noise with variance $\sigma^2$ applied to the $x$ and $y$ coordinates only, leaving $z$ unchanged. The \textit{perturbed global edge target point matrix} $\tilde{\mathbf{Q}}_e\in\mathbb{R}^{3\times n_e}$ was formed by replacing the indexed columns
\begin{align}
\tilde{\mathbf{Q}}_e(:,\mathcal{J}_{\mathrm{shoulder}})
=\tilde{\mathbf{Q}}_{e,\mathrm{shoulder}}, 
\label{eq:edge_replace_cols}
\end{align}
where 
\(
\tilde{\mathbf{Q}}_e(:,\{1,\ldots,n_e\}\!\setminus\!\mathcal{J}_{\mathrm{shoulder}})
=\mathbf{Q}_e(:,\{1,\ldots,n_e\}\!\setminus\!\mathcal{J}_{\mathrm{shoulder}})
\).
%
Finally, the local sewing line point matrix $\mathbf{Q}_s$ was obtained from the perturbed shoulder-edge points $\tilde{\mathbf{Q}}_{e,\mathrm{shoulder}}$ using \cite{soille1999morphological}, followed by smoothing the result using a Savitzky--Golay~\cite{5888646} filter (window length~$=11$, polynomial order~$=3$).


For validation, two alignment settings were considered:  
\begin{enumerate}[label=(\roman*)]
    \item \textbf{Edge $\tilde{\mathbf{Q}}_e$ + shoulder edge $\tilde{\mathbf{Q}}_{e,\mathrm{shoulder}}$ alignment:}
    Using $\tilde{\mathbf{Q}}_e$ and $\tilde{\mathbf{Q}}_{e,\mathrm{shoulder}}$ together with their corresponding CAD point matrices $\mathbf{P}_e$ and $\mathbf{P}_{e,\mathrm{shoulder}}$ as source point matrices, where $\mathbf{P}_{e,\mathrm{shoulder}}$ denotes the shoulder segment extracted from the CAD model.

    \item \textbf{Edge $\tilde{\mathbf{Q}}_e$ + sewing line $\mathbf{Q}_s$ alignment:}
    Using $\tilde{\mathbf{Q}}_e$, $\mathbf{Q}_s$, and their corresponding $\mathbf{P}_e$ and $\mathbf{P}_s$ as source point matrices.
\end{enumerate}


Table~\ref{tab:edge_vs_sewingline} shows that incorporating the sewing line $\mathbf{Q}_s$ improves alignment accuracy compared to using the perturbed shoulder edge points $\tilde{\mathbf{Q}}_{e,\mathrm{shoulder}}$. Although $\mathbf{Q}_s$ is approximately parallel to the perturbed edge points $\tilde{\mathbf{Q}}_e$, it provides complementary local constraints that compensate for edge perturbations from cutting deviations or surface occlusion. Consequently, the smoothed sewing line provides a stable reference for restoring the intended CAD geometry and improving pose-estimation precision.

\begin{table}[!h]
\centering
\caption{Comparison between the two alignment settings.}
\label{tab:edge_vs_sewingline}
\resizebox{\columnwidth}{!}{
\begin{tabular}{lcc}
\toprule
\multirow{2}{*}{\textbf{Method}} 
& \multicolumn{2}{c}{{$e_{\mathrm{Chamfer}}$ (mm)}} \\ 
\cmidrule(lr){2-3}
& {(Sew, U)} & {(Sew, O)} \\ 
\midrule
Edge $\tilde{\mathbf{Q}}_e$ + shoulder edge $\tilde{\mathbf{Q}}_{e,\mathrm{shoulder}}$
& $1.175 \pm0.094$ & $1.073\pm0.036$ \\[2pt]
Edge $\tilde{\mathbf{Q}}_e$ + sewing line $\mathbf{Q}_s$ 
& $\mathbf{0.892\pm0.232}$ & $\mathbf{0.808\pm0.053}$ \\
\bottomrule
\end{tabular}
}
\vspace{-3mm}
\end{table}
}

\subsection{Fabric alignment results}

\begin{figure*}[!t]
    \centering
    \includegraphics[width=1\textwidth]{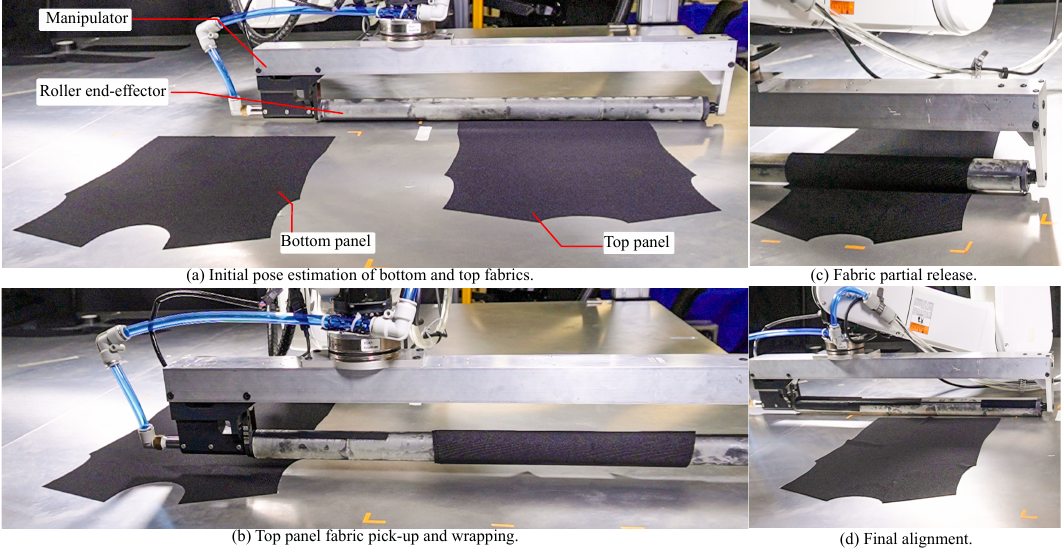}
    \caption{Fabric alignment experimental results.
    }
    \label{fig:exp_fabric_alignement}
    \vspace{-5mm}
\end{figure*}

To further evaluate the proposed fabric alignment system, real-world experiments with various fabric types (Section~\ref{sec:dataset}) were conducted following the full workflow described in Sec.~\ref{sec:step_1}--\ref{sec:step_4} under both unoccluded and partially occluded conditions (Fig.~\ref{fig:exp_fabric_alignement}).
After each trial, Chamfer RMSE (Eq.~\eqref{eq:e_chamfer}) between the top and bottom panel fabric edge points and sewing line points was measure for alignment accuracy.
Table~\ref{tab:comparison_summary_combined} shows that GLW-ICP achieves low Chamfer RMSE values across all tested fabric shapes, consistently reaching millimeter-level precision. These results demonstrate the method’s effectiveness and suitability for automated fabric alignment in real-world scenarios.

\subsection{Ablation Study and Adaptive Weighting Validation}

{\color{black}
    We conducted the ablation study comparing edge-only alignment, sewing-line-only alignment, and the proposed GLW-ICP based on both edge and sewing line information. As shown in Table~\ref{tab:ablation_three_way}, 1) Edge-only (global) alignment aligns the edge (global) well, but the alignment error of the sewing line (local) remains large; 2) Sew-only (local) alignment aligns the sewing line (local) well, but the alignment error of the edge (global) remains large; 3) The proposed GLW-ICP method keeps both sewing line (local) alignment error and edge (global) alignment error small. For the sewing operations, local sewing line alignment is necessary for accurate seaming, while global edge alignment of two fabric panels is required for the subsequent assembly processes. In this sense, the proposed GLW-ICP method demonstrates the best performance.
}

    \begin{table}[!b]
    \centering
    \footnotesize
    \setlength{\tabcolsep}{3pt}
    \setlength{\abovecaptionskip}{0pt}
    \setlength{\belowcaptionskip}{-4pt}
    \caption{Ablation study.}
    \label{tab:ablation_three_way}
    \begin{tabular}{@{}ll*{3}{>{\centering\arraybackslash}p{17mm}}@{}}
    \hline
    \multirow{3}{*}{\textbf{Fabric}} &
    \multirow{3}{*}{\textbf{Setting}} &
    \multicolumn{3}{>{\centering\arraybackslash}p{51mm}}{$e_{\mathrm{Chamfer}}$ (mm)} \\
    \cline{3-5}
    & & \multicolumn{2}{>{\centering\arraybackslash}p{34mm}}{\textbf{Unoccluded (U)}} & \multicolumn{1}{>{\centering\arraybackslash}p{17mm}}{\textbf{Occluded (O)}} \\
    \cline{3-5}
    & & {Edge} & {Sew} & {Sew} \\
    \hline
    \multirow{3}{*}{Collar}
    & Edge-only & 4.272$\pm$3.745 & 7.664$\pm$7.060 & 154.777$\pm$30.957  \\
    & Sew-only & 123.391$\pm$71.135 & 0.362$\pm$0.020 & 0.636$\pm$0.208  \\
    & GLW-ICP & 1.185$\pm$0.072 & 1.970$\pm$0.114 & 1.544$\pm$0.196 \\
    \hline
    \multirow{3}{*}{Rectangle}
    & Edge-only & 1.024$\pm$0.111 & 1.761$\pm$0.079 & 136.033$\pm$6.943  \\
    & Sew-only & 135.762$\pm$45.603 & 1.156$\pm$0.134 & 1.183$\pm$0.181  \\
    & GLW-ICP & 1.030$\pm$0.112 & 1.784$\pm$0.073 & 1.367$\pm$0.138  \\
    \hline
    \multirow{3}{*}{Shirt}
    & Edge-only & 0.775$\pm$0.052 & 0.782$\pm$0.073 & 167.049$\pm$1.086  \\
    & Sew-only & 2.280$\pm$0.140 & 0.890$\pm$0.050 & 1.198$\pm$0.166  \\
    & GLW-ICP & 0.777$\pm$0.052 & 0.791$\pm$0.077 & 0.720$\pm$0.035  \\
    \hline
    \end{tabular}
    \end{table}

{\color{black}
        We verify the weighting characteristic of GLW-ICP by two experiments on the shirt panel. In the first experiment, Gaussian noise with standard deviation $\sigma$ ranging from $0$ to $3$\,mm is added to the two shoulder parts, which account for about 25\% of the total number of edge points. $\bar{W}_e(\sigma)$ denotes the mean weight at noise level $\sigma$. The vertical axis in Fig.~\ref{fig:fig_edge_partial_meanweights_normalized} reports the ratio $\bar{W}_e(\sigma)/\bar{W}_e(\sigma{=}0)$, the mean weight normalized by its value at $\sigma{=}0$. As shown, the ratio of noised edge points $\bar{W}_{e_{\mathrm{noisy}}}^{U}$ decreases while that of non-noised edge points $\bar{W}_{e_{\mathrm{non-noisy}}}^{U}$ increases as $\sigma$ grows, following~\eqref{eq:W_e_1}. This demonstrates that GLW-ICP adjusts the weight of each edge point according to the noise injected into that point.
        

        In the second experiment, noise is applied to all edge points to show the behavior of the normalized mean weights between edge and sewing-line features. The normalized mean weights are defined as follows:
        \begin{equation}
        \begin{gathered}
        \bar{W}_e =
        \frac{\mathbf{1}_{n_e}^{\top}\mathbf{W}_e^{\top}\mathbf{1}_{n_e}/n_e}
        {\mathbf{1}_{n_e}^{\top}\mathbf{W}_e^{\top}\mathbf{1}_{n_e}/n_e + \mathbf{1}_{n_s}^{\top}\mathbf{W}_s^{\top}\mathbf{1}_{n_s}/n_s},\\[6pt]
        \bar{W}_s = 1-\bar{W}_e.
        \end{gathered}
        \label{eq:normalized_mean_weights}
        \end{equation}
        
        As shown in Fig.~\ref{fig:normalized_mean_weights}, as the edge noise standard deviation $\sigma$ increases, $\bar{W}_e$ increases, while $\bar{W}_s$ decreases accordingly. This result shows that the proposed mechanism achieves dynamic feature balance between the global edge feature and the local sewing line feature, thereby minimizing the overall loss.
        Finally, these two experiments demonstrate that GLW-ICP not only adjusts the mean weights of edge points according to the noise injected into the edge points, but also adjusts the normalized mean weights between edges and sewing lines.

        \begin{figure}[t]
        \centering
        \subfloat[Mean weights of edge points versus $\sigma$.\label{fig:fig_edge_partial_meanweights_normalized}]{%
            \includegraphics[width=0.85\columnwidth]{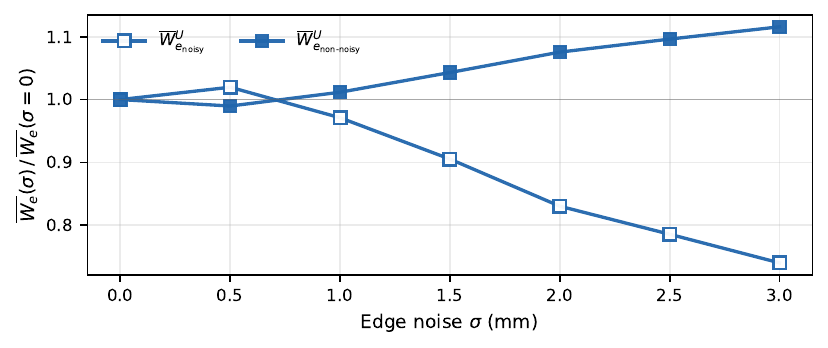}}\\[-2pt]
        \subfloat[Normalized mean weights versus $\sigma$.\label{fig:normalized_mean_weights}]{%
            \includegraphics[width=0.85\columnwidth]{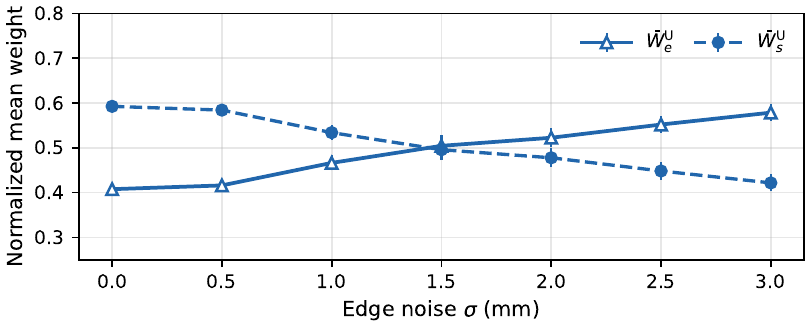}}
        \caption{Validation of adaptive weighting in GLW-ICP on the shirt panel under the unoccluded (U) setting. In (a), $\bar{W}_{e_{\mathrm{noisy}}}^{U}$ and $\bar{W}_{e_{\mathrm{non-noisy}}}^{U}$ denote the mean weights of the noised and non-noised edge points. In (b), $\bar{W}_e^U$ and $\bar{W}_s^U$ represent the normalized mean weights of edge and sewing line.}
        \label{fig:weight_validation}
        \end{figure}
    }

\acresetall
\section{Conclusions and future work}\label{sec:conclusion}
This paper presented a novel automated fabric alignment system that integrates GLW-ICP–based pose estimation and roller-based fabric manipulation. The proposed GLW-ICP algorithm jointly aligns edge and sewing line points under partial visibility using adaptive weighting and sparsity. Combined with a vision-guided robotic platform, including a 6-\ac{DoF} manipulator and suction roller end-effector, the system achieves accurate alignment through four main steps discussed in Sec. \ref{sec:sys_description}.
Experiments on real fabric panel shapes (collar, rectangle, and shirt) demonstrate millimeter-level precision and consistent performance under occlusion.

{\color{black}
    Beyond the current sewing-edge framework, the GLW-ICP method can be applied to more garment pose estimation cases. For instance, when aligning garments with rich surface features, visible garment contours can serve as global information, and printed logos, buttons, or stitching patterns can be used as local information under occlusion. Future work will explore the automatic selection of $k_{\text{ratio}}$ based on fabric shape and size. This will enable adaptation across different garment types and occlusion conditions without manual tuning. Additionally,  we plan to extend the system to handle wrinkled fabrics by integrating a template-mesh-based fabric state estimation method~\cite{tang2026rtffrandomtotargetfabricflattening} to flatten wrinkles before and during alignment.
}

\section*{Acknowledgment}
    The research work described in this paper was in part conducted in the JC STEM Lab of Robotics for Soft Materials, funded by The Hong Kong Jockey Club Charities Trust.

\bibliographystyle{IEEEtran}
\bibliography{bib/reference}

@article{ngan2011automated,
  title={Automated fabric defect detection—A review},
  author={Ngan, Henry YT and Pang, Grantham KH and Yung, Nelson HC},
  journal={Image and vision computing},
  volume={29},
  number={7},
  pages={442--458},
  year={2011},
  publisher={Elsevier}
}

@incollection{jana2015sewing,
  title={Sewing equipment and work aids},
  author={Jana, P},
  booktitle={Garment manufacturing technology},
  pages={275--315},
  year={2015},
  publisher={Elsevier}
}

@book{vilumsone2018industrial,
  title={Industrial cutting of textile materials},
  author={Vilumsone-Nemes, Ineta},
  year={2018},
  publisher={Woodhead Publishing}
}

@article{meng2022automatic,
  title={Automatic recognition of woven fabric structural parameters: a review},
  author={Meng, Shuo and Pan, Ruru and Gao, Weidong and Yan, Benchao and Peng, Yangyang},
  journal={Artificial Intelligence Review},
  volume={55},
  number={8},
  pages={6345--6387},
  year={2022},
  publisher={Springer}
}

@article{kim2023robotic,
  title={Robotic platform for automatic alignment and placement of fabric patterns for smart manufacturing in garment industry},
  author={Kim, Taehwan and Park, Yong-Lae},
  journal={International Journal of Precision Engineering and Manufacturing},
  volume={24},
  number={9},
  pages={1549--1561},
  year={2023},
  publisher={Springer}
}

@article{ku2023automated,
  title={Automated sewing system enabled by machine vision for smart garment manufacturing},
  author={Ku, Subyeong and Choi, HyunWoong and Kim, Ho-Young and Park, Yong-Lae},
  journal={IEEE Robotics and Automation Letters},
  year={2023},
  publisher={IEEE}
}

@article{parker1983robotic,
  title={Robotic fabric handling for automating garment manufacturing},
  author={Parker, JK and Dubey, R and Paul, FW and Becker, RJ},
  year={1983}
}

@INPROCEEDINGS{9551585,
  author={Shungo, Tajima and Hisashi, Date},
  booktitle={2021 IEEE 17th International Conference on Automation Science and Engineering (CASE)}, 
  title={Development of Fabric Feed Mechanism Using Horizontal Articulated Dual Manipulator for Automated Sewing}, 
  year={2021},
  volume={},
  number={},
  pages={1832-1837},
  doi={10.1109/CASE49439.2021.9551585}}

@INPROCEEDINGS{9419523,
  author={Manabe, K. and Tong, X. and Aiyama, Y.},
  booktitle={2021 IEEE International Conference on Intelligence and Safety for Robotics (ISR)}, 
  title={Single sheet separation method from piled fabrics using roller hand mechanism}, 
  year={2021},
  volume={},
  number={},
  pages={359-362},
  doi={10.1109/ISR50024.2021.9419523}}

@ARTICLE{9345958,
  author={Yamazaki, Kimitoshi and Abe, Taiki},
  journal={IEEE Robotics and Automation Letters}, 
  title={A Versatile End-Effector for Pick-and-Release of Fabric Parts}, 
  year={2021},
  volume={6},
  number={2},
  pages={1431-1438},
  doi={10.1109/LRA.2021.3057053}}

@article{kuvzel2022vacuum,
  title={Vacuum system for reinforcing fabric handling},
  author = {J. Ku\v{z}el and R. R\u{u}\v{z}ek},
  journal={Manufacturing Technology},
  volume={22},
  number={2},
  pages={204--210},
  year={2022}
}

@ARTICLE{atalie2020application,
  title={Application of digital CAD in whole-garment knitting manufacturing: A review},
  author={Atalie, DESALEGN},
  journal={International Journal of Science and Technology (STECH), AFRREV},
  volume={9},
  number={1},
  pages={1--17},
  year={2020}
}

@INPROCEEDINGS{ogura2022automation,
    author={Ogura, Eiri and Yoshimi, Takashi and Hirayama, Motoki},
    booktitle={2022 22nd International Conference on Control, Automation and Systems (ICCAS)}, 
    title={Automation of 3D sewing of two different shaped parts by robot arm - Evaluation and verification of the proposed method through experiments -}, 
    year={2022},
    volume={},
    number={},
    pages={179-183},
    doi={10.23919/ICCAS55662.2022.10003808}}

@inproceedings{schrimpf2012experiments,
  title={Experiments towards automated sewing with a multi-robot system},
  author={Schrimpf, Johannes and Wetterwald, Lars Erik},
  booktitle={2012 IEEE International Conference on Robotics and Automation},
  pages={5258--5263},
  year={2012},
  organization={IEEE}
}

@inproceedings{schrimpf2014velocity,
  title={Velocity coordination and corner matching in a multi-robot sewing cell},
  author={Schrimpf, Johannes and Bjerkeng, Magnus and Mathisen, Geir},
  booktitle={2014 IEEE/RSJ International Conference on Intelligent Robots and Systems},
  pages={4476--4481},
  year={2014},
  organization={IEEE}
}

@article{torgerson1988vision,
  title={Vision-guided robotic fabric manipulation for apparel manufacturing},
  author={Torgerson, Eric and Paul, Frank W},
  journal={IEEE Control Systems Magazine},
  volume={8},
  number={1},
  pages={14--20},
  year={1988},
  publisher={IEEE}
}

@ARTICLE{10197511,
  author={Ku, Subyeong and Choi, HyunWoong and Kim, Ho-Young and Park, Yong-Lae},
  journal={IEEE Robotics and Automation Letters}, 
  title={Automated Sewing System Enabled by Machine Vision for Smart Garment Manufacturing}, 
  year={2023},
  volume={8},
  number={9},
  pages={5680-5687},
  doi={10.1109/LRA.2023.3300284}}

@article{guo2022adaptive,
  title={Adaptive weighted robust iterative closest point},
  author={Guo, Yu and Zhao, Luting and Shi, Yan and Zhang, Xuetao and Du, Shaoyi and Wang, Fei},
  journal={Neurocomputing},
  volume={508},
  pages={225--241},
  year={2022},
  publisher={Elsevier}
}

@article{dang20203d,
  title={3D Registration for Self-Occluded Objects in Context},
  author={Dang, Zheng and Wang, Fei and Salzmann, Mathieu},
  journal={arXiv preprint arXiv:2011.11260},
  year={2020}
}

@inproceedings{ma2013robust,
  title={Robust estimation of nonrigid transformation for point set registration},
  author={Ma, Jiayi and Zhao, Ji and Tian, Jinwen and Tu, Zhuowen and Yuille, Alan L},
  booktitle={Proceedings of the IEEE conference on computer vision and pattern recognition},
  pages={2147--2154},
  year={2013}
}

@inproceedings{campbell2015adaptive,
  title={An adaptive data representation for robust point-set registration and merging},
  author={Campbell, Dylan and Petersson, Lars},
  booktitle={Proceedings of the IEEE international conference on computer vision},
  pages={4292--4300},
  year={2015}
}

@inproceedings{besl1992method,
  title={Method for registration of 3-D shapes},
  author={Besl, Paul J and McKay, Neil D},
  booktitle={Sensor fusion IV: control paradigms and data structures},
  volume={1611},
  pages={586--606},
  year={1992},
  organization={Spie}
}

@ARTICLE{4982554,
  author={Ying, Shihui and Peng, Jigen and Du, Shaoyi and Qiao, Hong},
  journal={IEEE Transactions on Automation Science and Engineering}, 
  title={A Scale Stretch Method Based on ICP for 3D Data Registration}, 
  year={2009},
  volume={6},
  number={3},
  pages={559-565},
  doi={10.1109/TASE.2009.2021337}}

@ARTICLE{1430845,
  author={Yonghuai Liu},
  journal={IEEE Transactions on Systems, Man, and Cybernetics, Part B (Cybernetics)}, 
  title={Eliminating false matches for the projective registration of free-form surfaces with small translational motions}, 
  year={2005},
  volume={35},
  number={3},
  pages={607-624},
  doi={10.1109/TSMCB.2005.843978}}

@article{bearee2011innovative,
  title={An innovative subdivision-ICP registration method for tool-path correction applied to deformed aircraft parts machining},
  author={B{\'e}ar{\'e}e, Richard and Dieulot, Jean-Yves and Rabat{\'e}, Patrice},
  journal={The International Journal of Advanced Manufacturing Technology},
  volume={53},
  pages={463--471},
  year={2011},
  publisher={Springer}
}

@inproceedings{rusinkiewicz2001efficient,
  title={Efficient variants of the ICP algorithm},
  author={Rusinkiewicz, Szymon and Levoy, Marc},
  booktitle={Proceedings third international conference on 3-D digital imaging and modeling},
  pages={145--152},
  year={2001},
  organization={IEEE}
}

@article{zhang2002flexible,
  title={A flexible new technique for camera calibration},
  author={Zhang, Zhengyou},
  journal={IEEE Transactions on pattern analysis and machine intelligence},
  volume={22},
  number={11},
  pages={1330--1334},
  year={2002},
  publisher={IEEE}
}

@article{kobayashi2025rollup,
  title={RollUP: Rolling-up End-effector for Fabric Handing},
  author={Kobayashi, Akinari and Dong, Wenbo and Seino, Akira and Tokuda, Fuyuki and Kosuge, Kazuhiro},
  journal={Authorea Preprints},
  publisher={Authorea}
}

@INPROCEEDINGS{924423,
  author={Rusinkiewicz, S. and Levoy, M.},
  booktitle={Proceedings Third International Conference on 3-D Digital Imaging and Modeling}, 
  title={Efficient variants of the ICP algorithm}, 
  year={2001},
  volume={},
  number={},
  pages={145-152},
  doi={10.1109/IM.2001.924423}}

@book{soille1999morphological,
  title={Morphological image analysis: principles and applications},
  author={Soille, Pierre and others},
  volume={2},
  number={3},
  year={1999},
  publisher={Springer}
}

@ARTICLE{11091470,
  author={Tang, Kai and Huang, Xuzhao and Seino, Akira and Tokuda, Fuyuki and Kobayashi, Akinari and Tien, Norman C. and Kosuge, Kazuhiro},
  journal={IEEE Robotics and Automation Letters}, 
  title={Fixture-Free Automated Sewing System Using Dual-Arm Manipulator and High-Speed Fabric Edge Detection}, 
  year={2025},
  volume={10},
  number={9},
  pages={8962-8969},
  }

@ARTICLE{10715572,
  author={Tokuda, Fuyuki and Murakami, Ryo and Seino, Akira and Kobayashi, Akinari and Hayashibe, Mitsuhiro and Kosuge, Kazuhiro},
  journal={IEEE Transactions on Automation Science and Engineering}, 
  title={Fixture-Free 2D Sewing Using a Dual-Arm Manipulator System}, 
  year={2025},
  volume={22},
  number={},
  pages={7927-7940},
  doi={10.1109/TASE.2024.3474914}}

@misc{tang2026rtffrandomtotargetfabricflattening,
      title={RTFF: Random-to-Target Fabric Flattening Policy using Dual-Arm Manipulator}, 
      author={Kai Tang and Dipankar Bhattacharya and Hang Xu and Fuyuki Tokuda and Norman C. Tien and Kazuhiro Kosuge},
      year={2026},
      eprint={2510.00814},
      archivePrefix={arXiv},
      primaryClass={cs.RO},
      url={https://arxiv.org/abs/2510.00814}, 
}

@inproceedings{chetverikov2002trimmed,
  title={The trimmed iterative closest point algorithm},
  author={Chetverikov, Dmitry and Svirko, Dmitry and Stepanov, Dmitry and Krsek, Pavel},
  booktitle={2002 International Conference on Pattern Recognition},
  volume={3},
  pages={545--548},
  year={2002},
  organization={IEEE}
}

@inproceedings{bouaziz2013sparse,
  title={Sparse iterative closest point},
  author={Bouaziz, Sofien and Tagliasacchi, Andrea and Pauly, Mark},
  booktitle={Computer graphics forum},
  volume={32},
  number={5},
  pages={113--123},
  year={2013},
  organization={Wiley Online Library}
}

@article{myronenko2010point,
  title={Point set registration: Coherent point drift},
  author={Myronenko, Andriy and Song, Xubo},
  journal={IEEE transactions on pattern analysis and machine intelligence},
  volume={32},
  number={12},
  pages={2262--2275},
  year={2010},
  publisher={IEEE}
}

@ARTICLE{9336308,
  author={Zhang, Juyong and Yao, Yuxin and Deng, Bailin},
  journal={IEEE Transactions on Pattern Analysis and Machine Intelligence}, 
  title={Fast and Robust Iterative Closest Point}, 
  year={2022},
  volume={44},
  number={7},
  pages={3450-3466},
  doi={10.1109/TPAMI.2021.3054619}}

@ARTICLE{10061449,
  author={Lv, Chenlei and Lin, Weisi and Zhao, Baoquan},
  journal={IEEE Transactions on Image Processing}, 
  title={KSS-ICP: Point Cloud Registration Based on Kendall Shape Space}, 
  year={2023},
  volume={32},
  number={},
  pages={1681-1693},
  doi={10.1109/TIP.2023.3251021}}

@ARTICLE{5888646,
  author={Schafer, Ronald W.},
  journal={IEEE Signal Processing Magazine}, 
  title={What Is a Savitzky-Golay Filter? [Lecture Notes]}, 
  year={2011},
  volume={28},
  number={4},
  pages={111-117},
  doi={10.1109/MSP.2011.941097}}

\begin{IEEEbiography}[{\includegraphics[width=1in,height=1.25in,clip,keepaspectratio]
{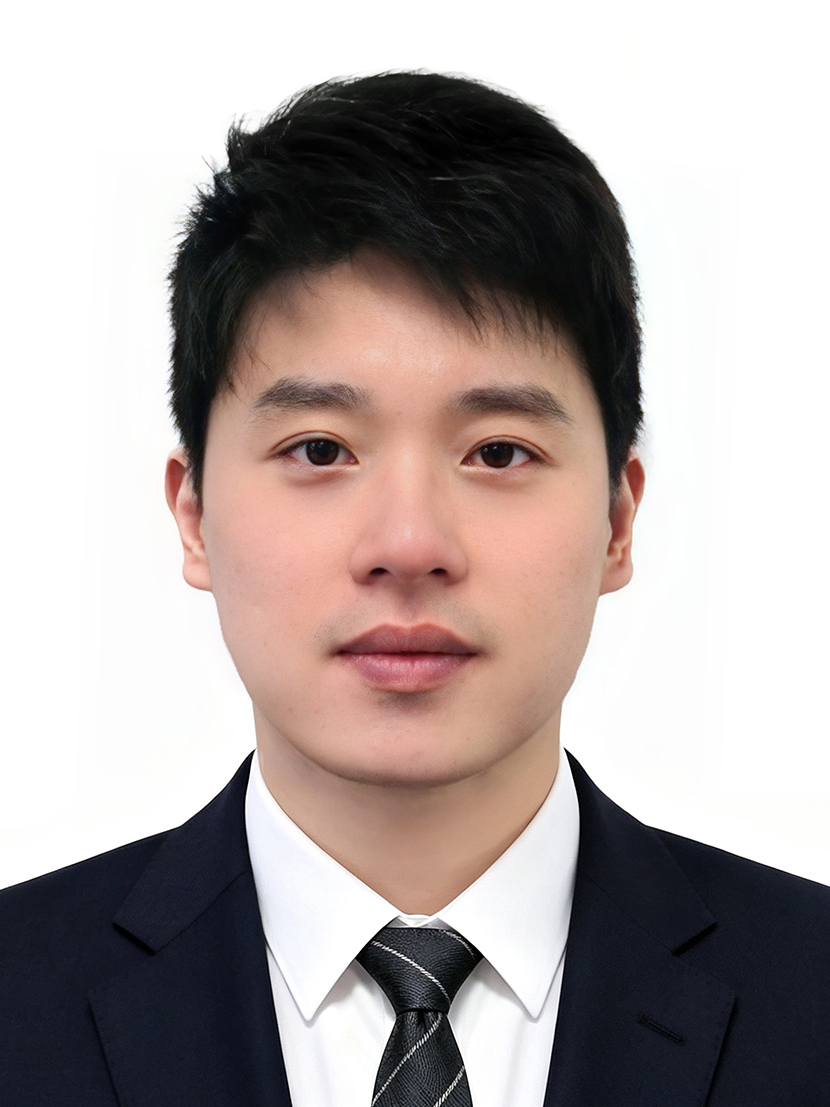}}]{Wenbo Dong }
(Graduate Student Member, IEEE) received the B.Sc. degree in Automation from Northeastern University, China, the M.Sc. degree in Control Engineering from the Harbin Institute of Technology, China, in 2017, and the M.Sc. degree in Mechanical Engineering from the University of California, Riverside, USA, in 2022. He is currently pursuing the Ph.D. degree in robotics with the JC STEM Lab of Robotics for Soft Materials, The University of Hong Kong, Hong Kong SAR. From 2017 to 2021, he was a Research Assistant with the Shenyang Institute of Automation, Chinese Academy of Sciences, China. From 2022 to 2025, he was with the Centre for Transformative Garment Production, Hong Kong SAR, in collaboration with Tohoku University, Japan. His research interests include robotic manipulation and perception.
\end{IEEEbiography}

\begin{IEEEbiography}[{\includegraphics[width=1in,height=1.25in,clip,keepaspectratio]
{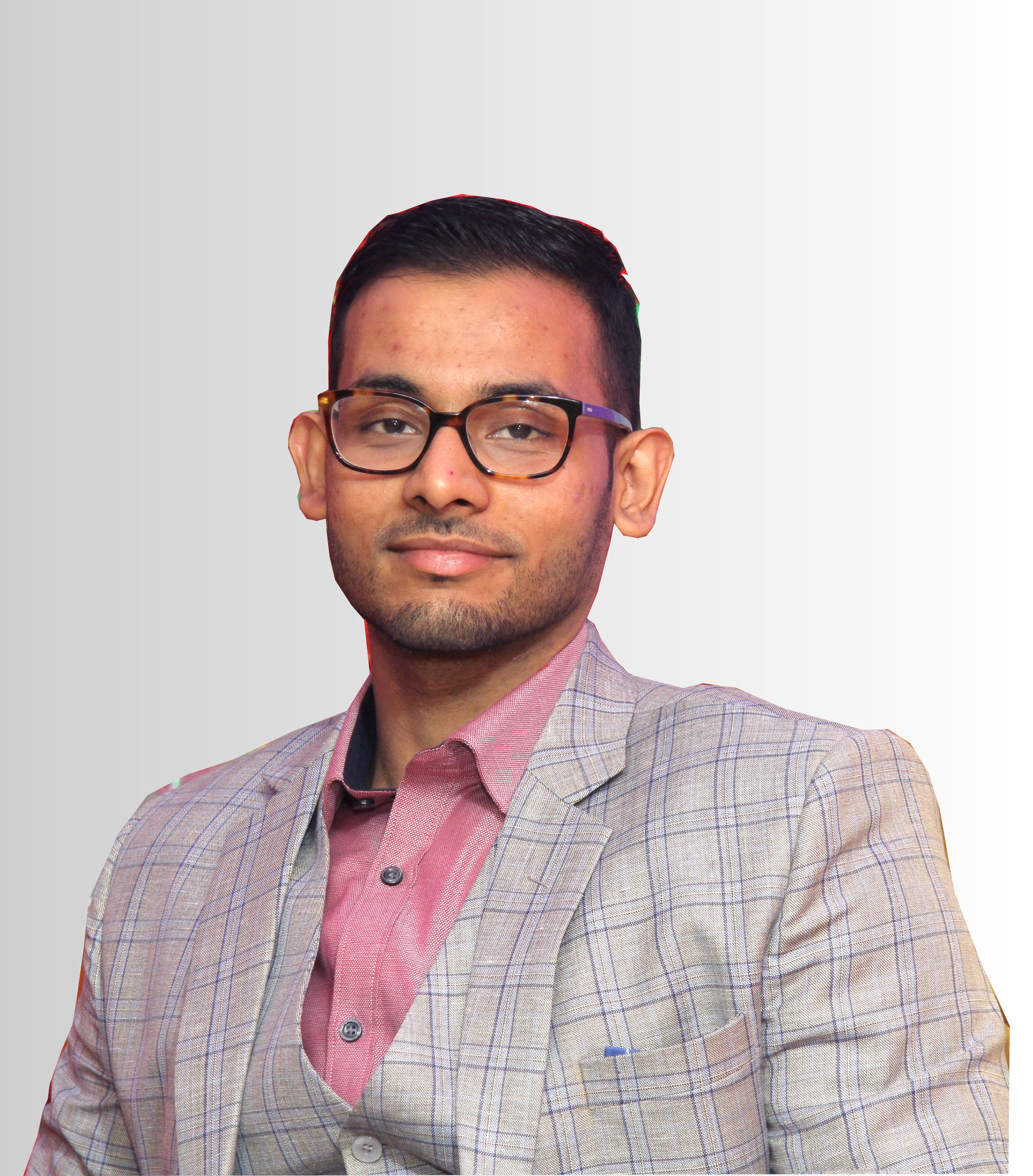}}]{Dipankar Bhattacharya }
(Member, IEEE) received the B.Tech. degree in electronics and communication engineering from NERIST, India, in 2010, the M.Tech. degree in systems and control from IIT Roorkee, India, in 2013, and the Ph.D. degree in mechatronics engineering from The University of Auckland, New Zealand, in 2021. From 2024 to 2025, he was a Senior Research Engineer with the Center for Transformative Garment Production (TransGP), Hong Kong, where he developed learning-based robotic systems for fabric manipulation, alignment, and automated sewing. Since 2025, he has been a Marie Skłodowska-Curie Fellow with the Dyson School of Design Engineering, Imperial College London, London, U.K. His research interests include assistive and rehabilitation robotics, learning-based control, deformable-object manipulation, and cable-driven robots.

\end{IEEEbiography}

\begin{IEEEbiography}
[{\includegraphics[width=1in,height=1.25in,clip,keepaspectratio]
{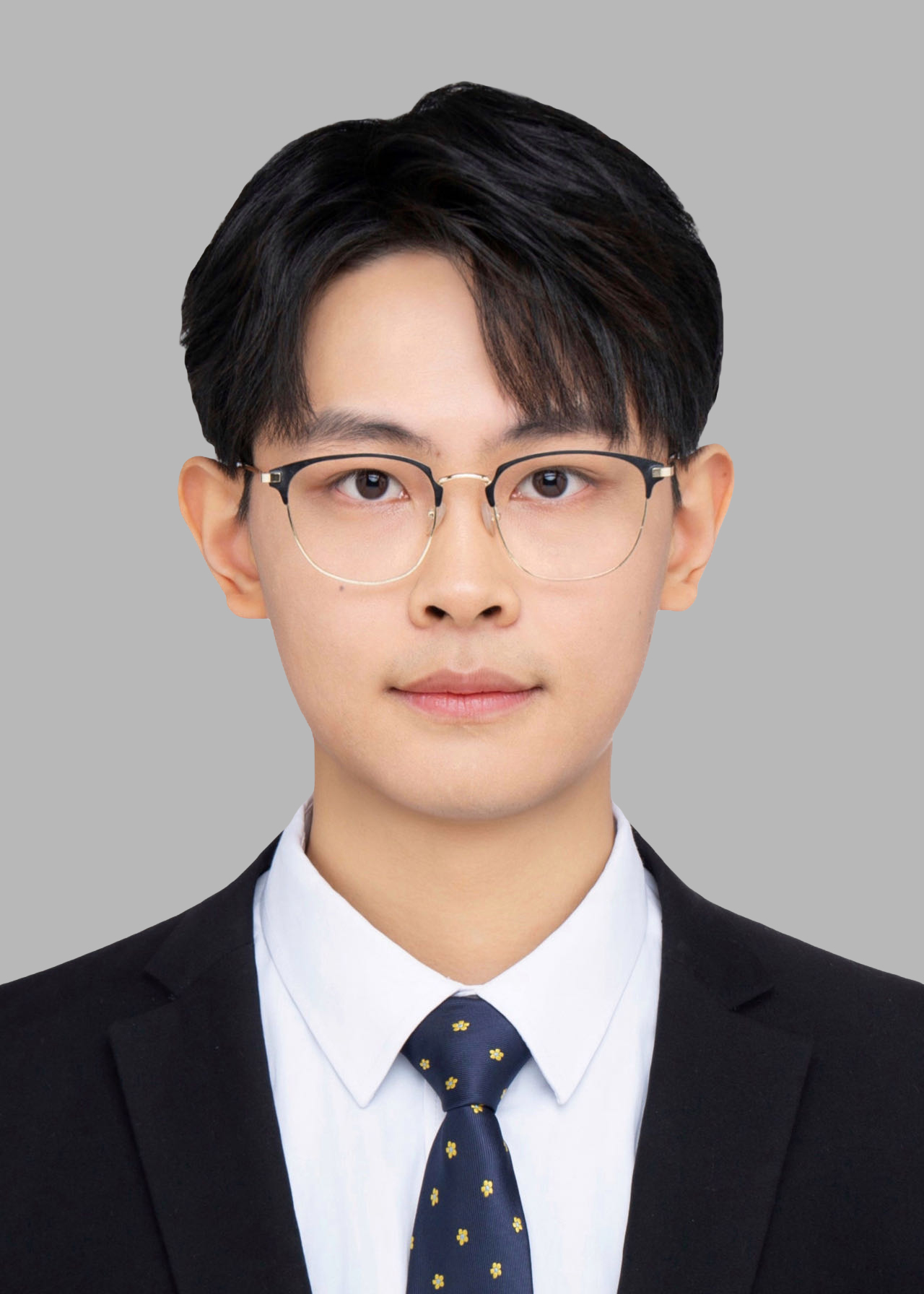}}]{Kai Tang} (Graduate Student Member, IEEE) received his B.Eng. degree in Process Equipment and Control Engineering from South China University of Technology in 2020, and M.Sc. degree with Distinction in Control Systems from Imperial College London in 2021. He is currently pursuing his Ph.D. degree in robotics at the JC STEM Lab of Robotics for Soft Materials, The University of Hong Kong. From 2022 to 2025, he worked with the Centre for Transformative Garment Production, Hong Kong SAR, which was in collaboration with Tohoku University, Japan. His research focuses on robot learning and control for fabric manipulation and fixture-free automated sewing.
\end{IEEEbiography}

\begin{IEEEbiography}[{\includegraphics[width=1in,height=1.25in,clip,keepaspectratio]
{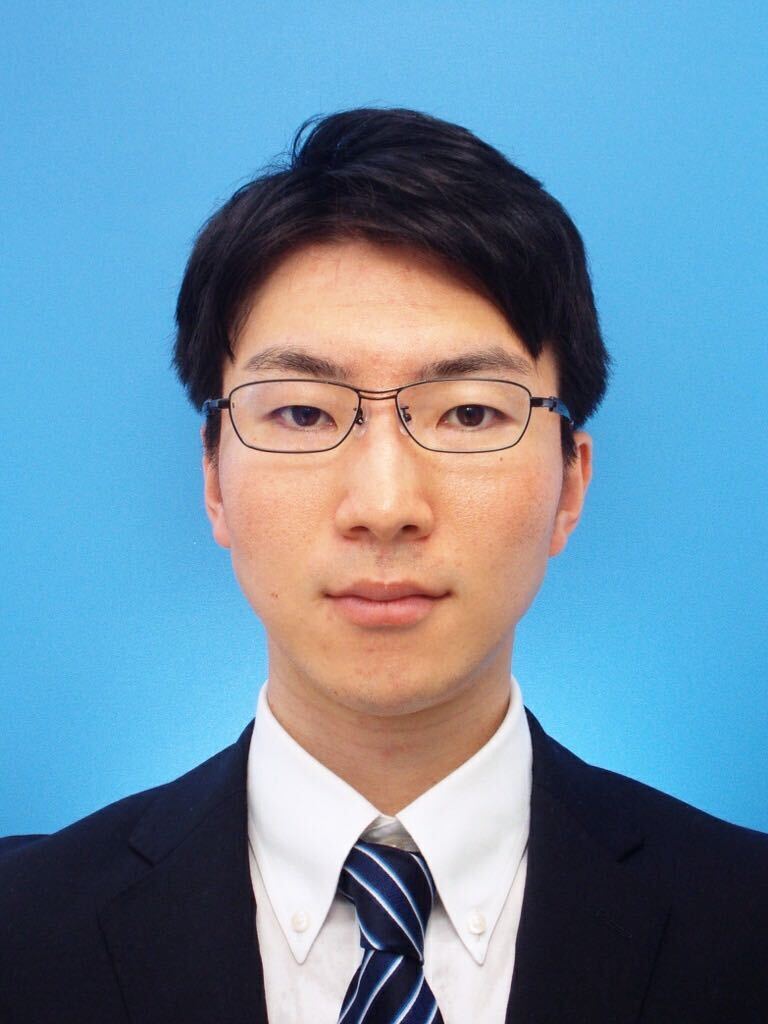}}]{Akinari Kobayashi }
(Member, IEEE) received
the B.S. and M.S. degree in engineering from
Tohoku University, Sendai, Japan, in 2013 and
2017, respectively. He received the Ph.D. de-
gree in engineering from Tohoku University,
Sendai, Japan, in 2020.
He was a Research Officer with the
Centre for Transformative Garment Production,
Hong Kong from 2021 to 2025, and a Visiting Research
Associate with The University of Hong Kong,
Hong Kong SAR from 2022 to 2025. His research focuses on robot hands, robotic manipulation, and robot sewing.
\end{IEEEbiography}

\begin{IEEEbiography}[{\includegraphics[width=1in,height=1.25in,clip,keepaspectratio]
{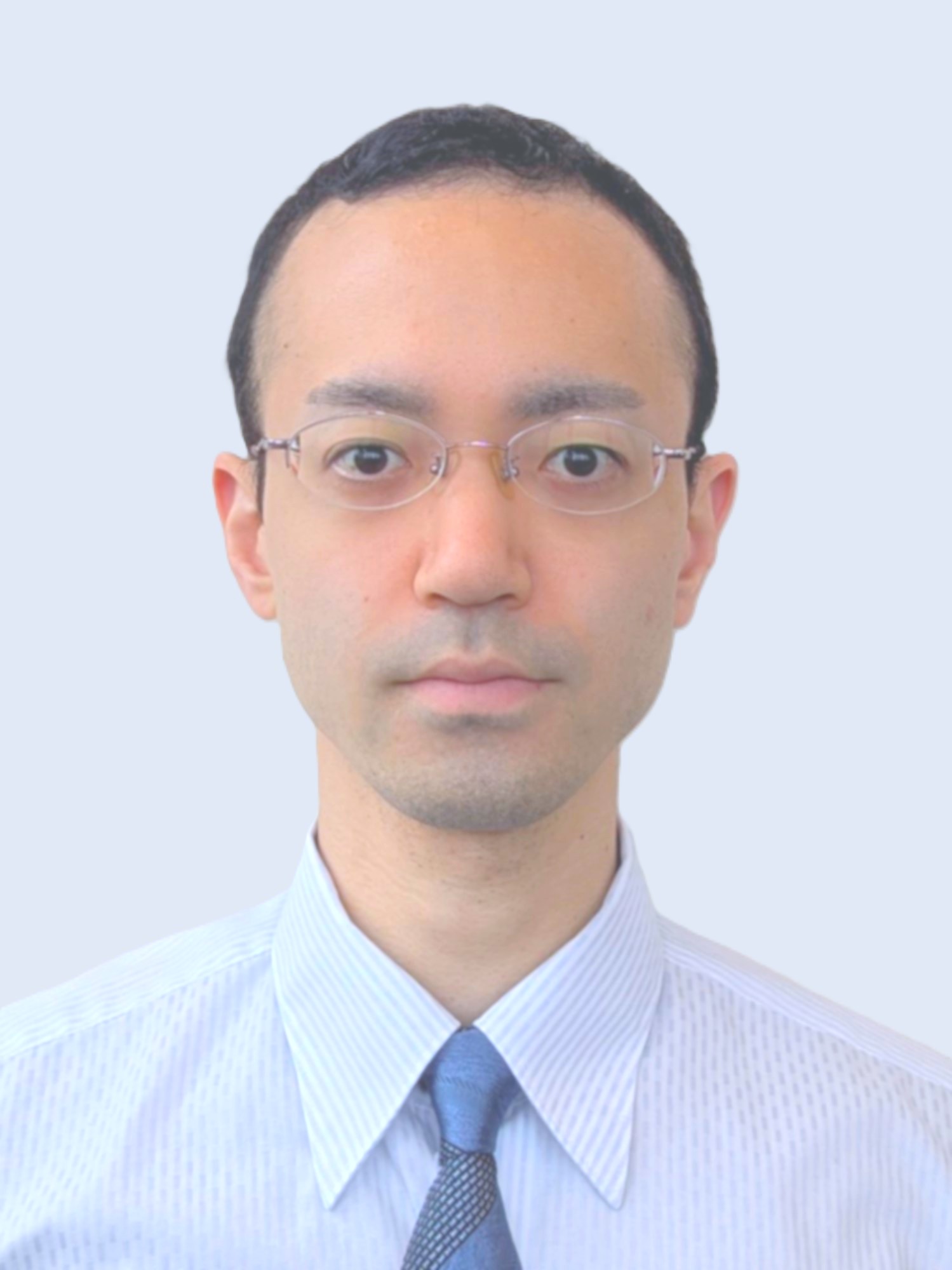}}]{Fuyuki Tokuda }
(Member, IEEE) received the B.S. degree in engineering from the Nagoya Institute of Technology, Nagoya, Japan, in 2017, and the M.S. and Ph.D. degrees in engineering from Tohoku University, Sendai, Japan, in 2019 and 2022, respectively. From 2022 to 2023, he was a Postdoctoral Fellow with the Centre for Transformative Garment Production (TransGP), an InnoHK Research Centre jointly established by The University of Hong Kong and Tohoku University, Hong Kong. From 2023 to 2025, he was a Research Officer with TransGP and concurrently was a Visiting Research Associate with The University of Hong Kong from 2022 to 2025. Since 2025, he has been an Assistant Professor with the Unprecedented-Scale Data Analytics Center, Tohoku University, and is also affiliated with the Graduate School of Information Sciences, Tohoku University. Dr. Tokuda was the recipient of the Research Fellowship from the Tohoku University Graduate Program for Integration of Mechanical Systems in 2018 and Research Fellowship from the Japan Society for the Promotion of Science in 2021.
\end{IEEEbiography}

\begin{IEEEbiography}[{\includegraphics[width=1in,height=1.25in,clip,keepaspectratio]
{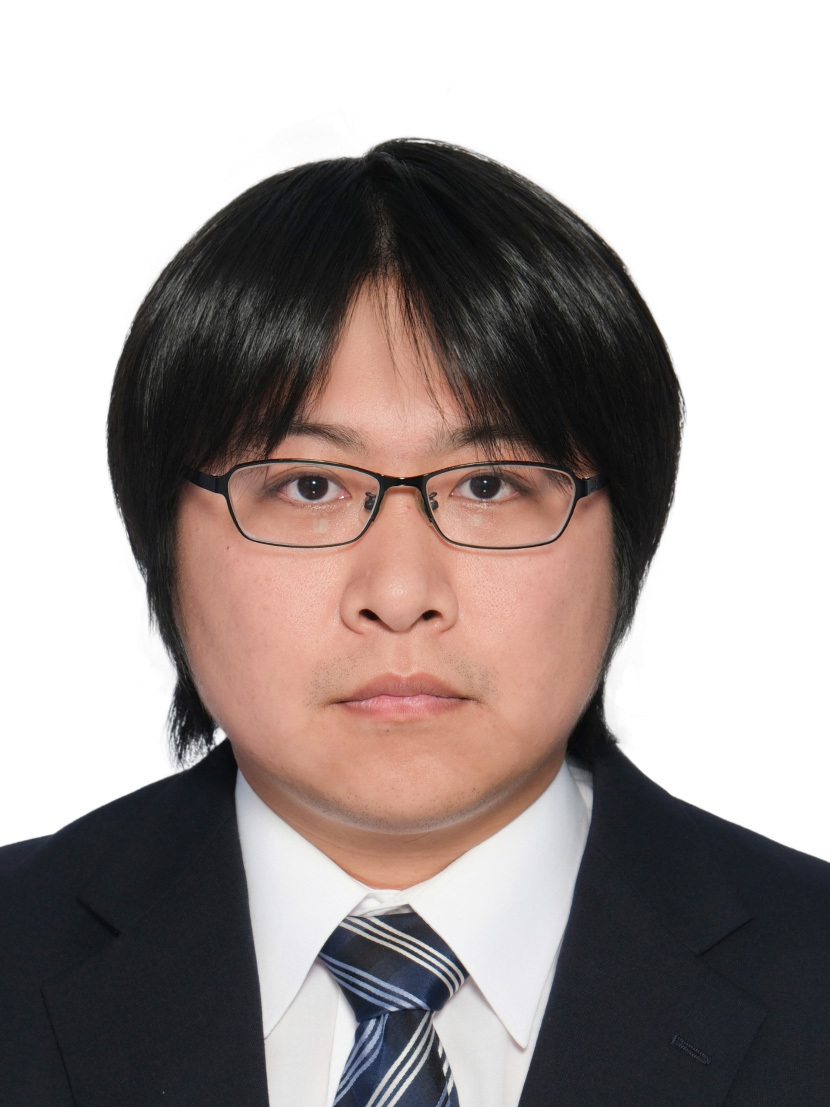}}]{Akira Seino }
(Member, IEEE) received the B.S. degree in Engineering from the Department of the Mechanical System Engineering at Yamagata University, Japan, in 2014, and the M.S. and Ph.D. degrees in Engineering from the Department of Bioengineering and Robotics, and the Department of Robotics at Tohoku University, Japan, in 2016 and 2019, respectively. 
From 2019 to 2021, he was a Project Assistant Professor in the Faculty of Symbiotic Systems Science at Fukushima University, Japan. 
He was a Research Fellow at Transformative AI \& Robotics International Research Center at Tohoku University, Japan, in 2021. 
He is a Research Officer at Centre for Transformative Garment Production, Hong Kong SAR, and a Visiting Research Associate in the Department of Electrical and Electronic Engineering at The University of Hong Kong, Hong Kong SAR. 
His research interests include mechanical design for robots, robot technology for industrial applications, and control of power-assisted systems. 
He is a member of the IEEE Robotics and Automation Society (IEEE RAS).
\end{IEEEbiography}

\begin{IEEEbiography}[{\includegraphics[width=1in,height=1.25in,clip,keepaspectratio]
{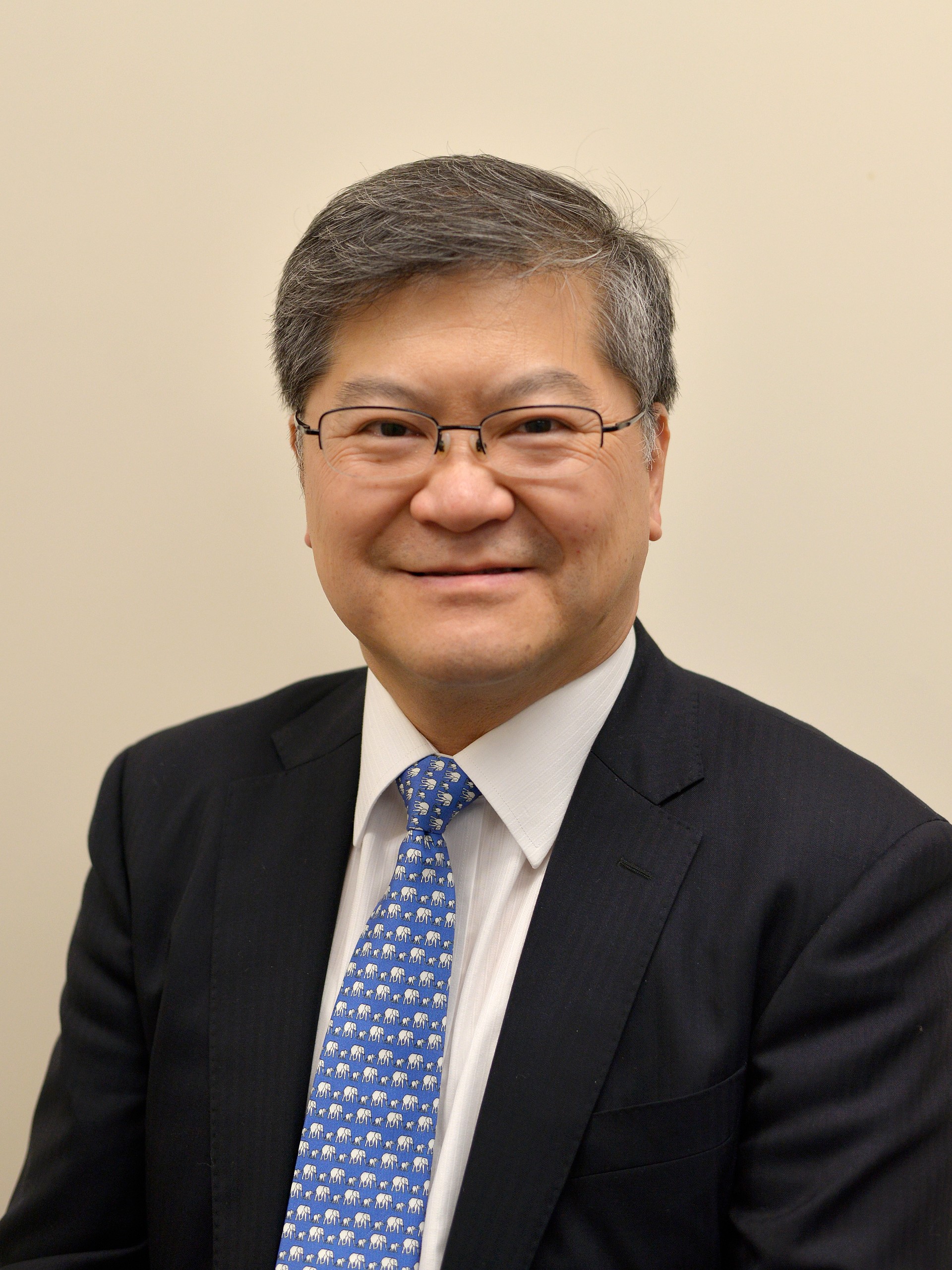}}]{Norman C. Tien  }
received the B.S. degree in
engineering physics from the University of California at Berkeley, Berkeley, CA, USA, in 1981,
the M.S. degree in electrical engineering from
the University of Illinois, Champaign, IL, USA, in
1984, and the Ph.D. degree in electrical engineering from the University of California at San
Diego, La Jolla, CA, USA, in 1993.

He is currently the Taikoo Professor of Engineering and Chair Professor of Microsystems
Technology with The University of Hong Kong
(HKU), Hong Kong. He is also the Head of Innovation Academy of
Faculty of Engineering and the Managing Director of the Centre for
Transformative Garment Production. He was the Dean of Engineering
from 2012 to 2018, and was the Vice-President and Pro-Vice-Chancellor
(Institutional Advancement) from 2019 to 2021 with HKU. Prior to joining
HKU, he was the Nord Professor of Engineering with Case Western
Reserve University, Cleveland, OH, USA, where he was the Dean of
Engineering from 2007 to 2011. He previously held faculty positions
with the University of California at Davis, Davis, CA, USA, University of
California at Berkeley, Berkeley, CA, USA, and Cornell University, Ithaca,
NY, USA. 
\end{IEEEbiography}

\begin{IEEEbiography}[{\includegraphics[width=1in,height=1.25in,clip,keepaspectratio]
{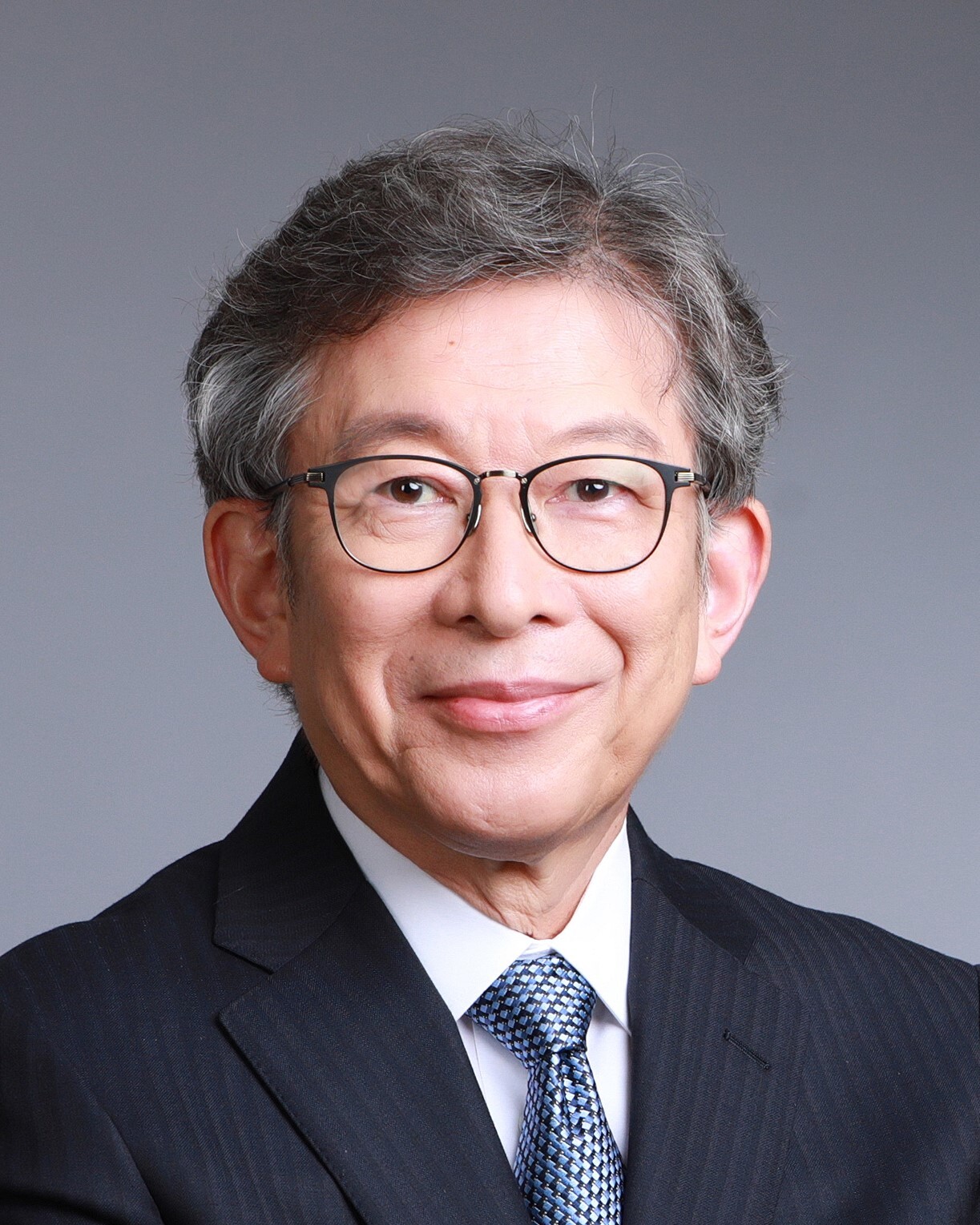}}]{Kazuhiro Kosuge}
(Life Fellow, IEEE) is Chair Professor of Robotic Systems in the Department of Mechanical Engineering at the City University of Hong Kong. He received his B.Sc., M.Sc., and Ph.D. in Control Engineering from the Tokyo Institute of Technology in 1978, 1980, and 1988, respectively. Following positions as an R\&D Staff member in the Production Engineering Department at Denso Corporation and Research Associate at the Tokyo Institute of Technology, he joined Nagoya University as Associate Professor and subsequently Tohoku University as Professor in 1995, where he was appointed Distinguished Professor in 2018. From 2021, he served as Chair Professor of Robotic Systems at the University of Hong Kong before joining the City University of Hong Kong in June 2026. 

In 2018, he was awarded the Medal of Honor with Purple Ribbon by the Japanese Government in recognition of his outstanding contributions to academic and industrial advancement. In 2021, he received the IEEE RAS George Saridis Leadership Award in Robotics and Automation in recognition of his innovative research vision and exceptional leadership in the robotics and automation community. He is an IEEE Life Fellow, a Fellow of JSME, SICE, RSJ, and JSAE, and a Member of the Engineering Academy of Japan. His service to the profession includes serving as President of the IEEE Robotics and Automation Society (2010–2011), IEEE Division X Director (2015–2016), and IEEE Vice President for Technical Activities (2020).

\end{IEEEbiography}

\end{document}